\documentclass[10pt,journal,compsoc]{IEEEtran}

\ifCLASSOPTIONcompsoc
   \usepackage[nocompress]{cite}
\else
\fi
\usepackage{url}
\usepackage{graphicx}
\usepackage{amsmath,amssymb} 
\usepackage{color}
\usepackage{subfigure}

\usepackage{bm}
\usepackage{color}
\newcommand{\x} {\textbf{x}}

\newcommand{\y} {\textbf{y}}
\newcommand{\w} {\textbf{w}}
\newcommand{\W} {\textbf{W}}

\newcommand{\I} {\textbf{I}}

\newcommand{\K} {\textbf{K}}
\newcommand{\kk} {\textbf{k}}
\newcommand{\Y} {\textbf{Y}}
\newcommand{\LL} {\textbf{L}}

\usepackage{algorithmic}
\usepackage{multirow}

\DeclareMathAlphabet{\mathpzc}{OT1}{pzc}{m}{it} %
\DeclareMathOperator*{\argmax}{argmax}
\DeclareMathOperator*{\argmin}{argmin}

\newcommand{\trans}[1]{{#1}^{\ensuremath{\mathsf{T}}}} 

\newcommand{\etal}{\emph{et al.}}
\newcommand{\eg}{\emph{e.g.}}
\newcommand{\ie}{\emph{i.e.}~}
\begin{document}
%
\title{Learning Deep Representation for Face Alignment with Auxiliary Attributes}

\author{Zhanpeng~Zhang,
        Ping~Luo,
        Chen~Change~Loy,~\IEEEmembership{Member,~IEEE}
        and~Xiaoou~Tang,~\IEEEmembership{Fellow,~IEEE}
       \IEEEcompsocitemizethanks{\IEEEcompsocthanksitem The authors are with the Department of Information Engineering, The Chinese University of Hong Kong, Hong Kong. \protect\\
       E-mail: \{zz013, pluo, ccloy, xtang\}@ie.cuhk.edu.hk\protect\\
}
}

\markboth{to appear in IEEE Transactions on Pattern Analysis and Machine Intelligence}%
{Shell \MakeLowercase{\textit{et al.}}: Bare Demo of IEEEtran.cls for Computer Society Journals}

\IEEEtitleabstractindextext{%
\begin{abstract}
In this study, we show that landmark detection or face alignment task is not a single and independent problem. Instead, its robustness can be greatly improved with auxiliary information. Specifically, we jointly optimize landmark detection together with the recognition of heterogeneous but subtly correlated facial attributes, such as gender, expression, and appearance attributes. This is non-trivial since different attribute inference tasks have different learning difficulties and convergence rates. To address this problem, we formulate a novel tasks-constrained deep model, which not only learns the inter-task correlation but also employs dynamic task coefficients to facilitate the optimization convergence when learning multiple complex tasks. Extensive evaluations show that the proposed task-constrained learning (i) outperforms existing face alignment methods, especially in dealing with faces with severe occlusion and pose variation, and (ii) reduces model complexity drastically compared to the state-of-the-art methods based on cascaded deep model.
\end{abstract}

}

\maketitle
\IEEEpeerreviewmaketitle

\section{Introduction}
\label{sec:introduction}
\begin{figure*}[t]
\centering
\subfigure[]{
 \label{fig:highlighta} 
\includegraphics[width=3.75in]{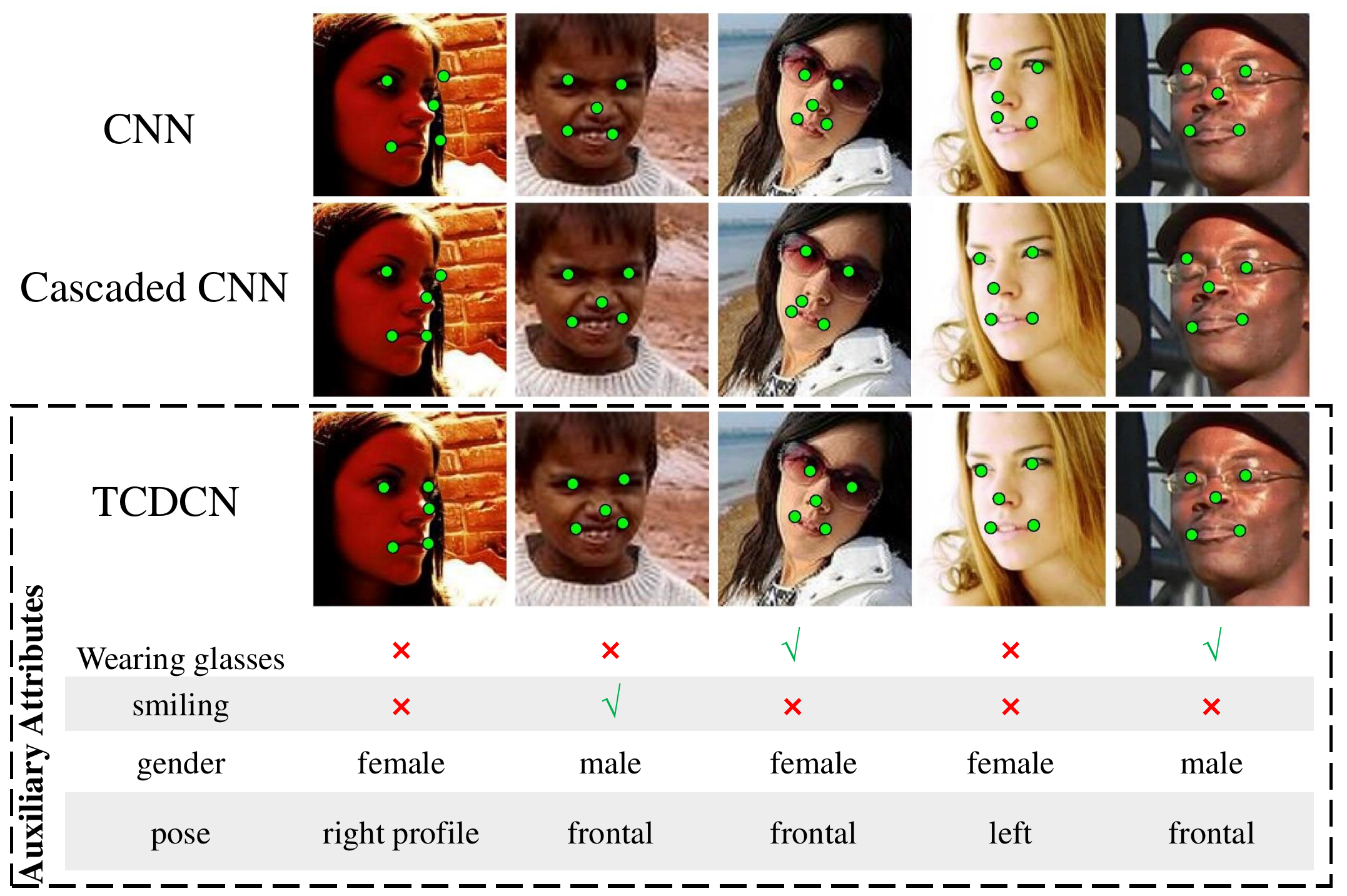}}
\subfigure[]{
\label{fig:highlightb} 
\includegraphics[width=3in]{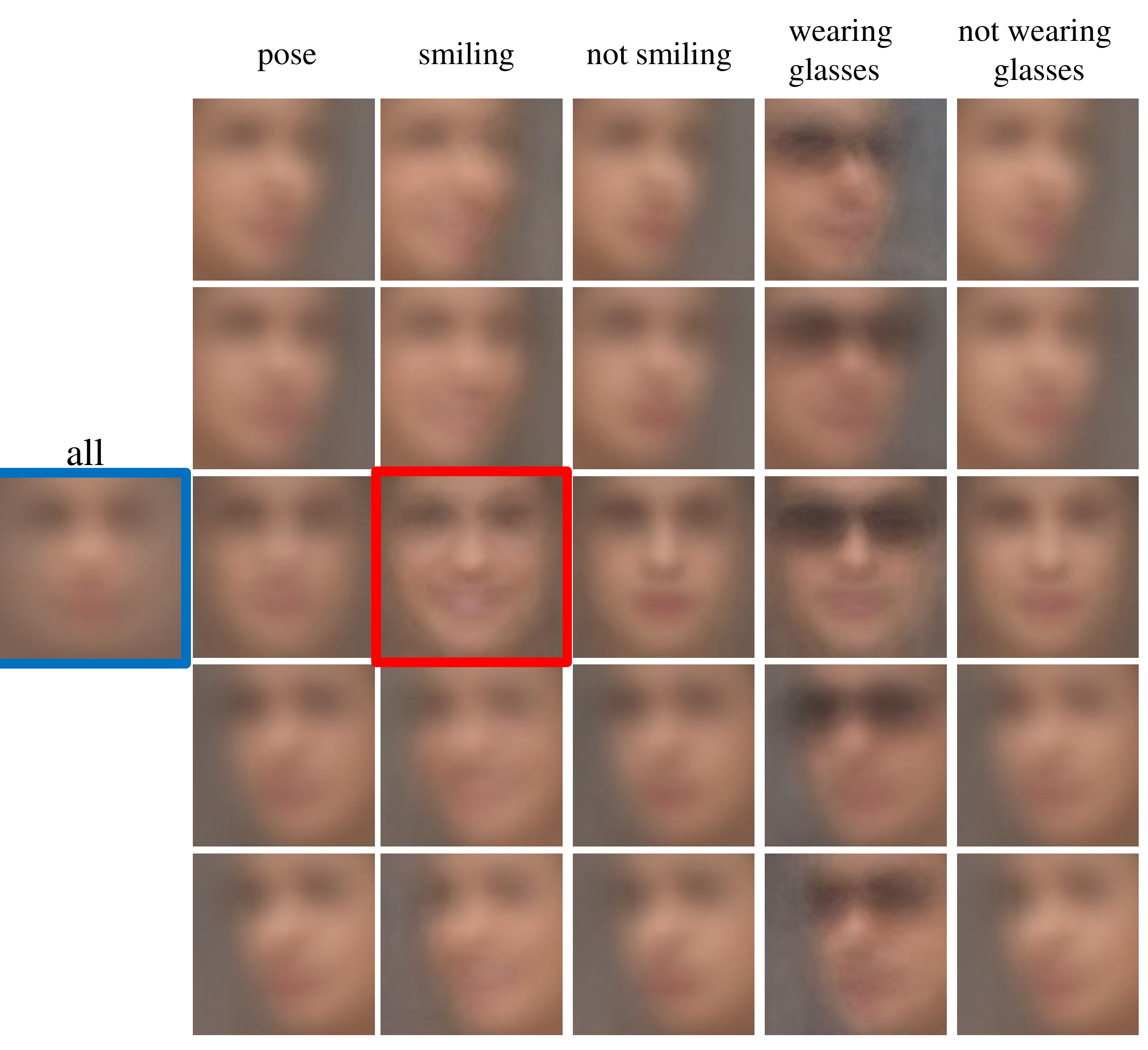}}
\caption{(a) Examples of facial landmark detection by a single conventional CNN, the cascaded CNN~\cite{Sun2013}, and the proposed Tasks-Constrained Deep Convolutional Network (TCDCN). More accurate detection can be achieved by optimizing the detection task jointly with related/auxiliary tasks. (b) Average face images with different attributes. The image in blue rectangle is averaged among the whole training faces, while the one in red is from the smiling faces with frontal pose. It indicates that the input and solution space can be effectively divided into subsets, which are in different distributions. This lowers the learning difficulty.}

\label{fig:highlight} 
\end{figure*}

Face alignment, or detecting semantic facial landmarks (\eg, eyes, nose, mouth corners) is a fundamental component in many face analysis tasks, such as facial attribute inference~\cite{kumarattribute}, face verification~\cite{LuT14}, and face recognition~\cite{huang2013coupling}. Though great strides have been made in this field (see Sec.~\ref{sec:related_work}), robust facial landmark detection remains a formidable challenge in the presence of partial occlusion and large head pose variations (Fig.~\ref{fig:highlight}).

Landmark detection is traditionally approached as a single and independent problem. Popular approaches include template fitting approaches~\cite{Cootes2001,Zhu2012,Yu2013,tzimiropoulos2014gauss} and regression-based methods~\cite{dollar13,Cao2012,Cootes2012,yang2013sieving,zhang2014coarse}. More recently, deep models have been applied too. For example, Sun~\etal~\cite{Sun2013} propose to detect facial landmarks by coarse-to-fine regression using a cascade of deep convolutional neural networks (CNN). This method shows superior accuracy compared to previous methods~\cite{Belhumeur2011,Cao2012} and existing commercial systems. Nevertheless, the method requires a complex and unwieldy cascade architecture of deep model.

We believe that facial landmark detection is not a standalone problem, but its estimation can be influenced by a number of heterogeneous and subtly correlated factors. Changes on a face are often governed by the same rules determined by the intrinsic facial structure.
%
%
For instance, when a kid is smiling, his mouth is widely opened (the second image in Fig.~\ref{fig:highlighta}). Effectively discovering and exploiting such an intrinsically correlated facial attribute would help in detecting the mouth corners more accurately. Also, the inter-ocular distance is smaller in faces with large yaw rotation (the first image in Fig.~\ref{fig:highlighta}). Such pose information can be leveraged as an additional source of information to constrain the solution space of landmark estimation.
Indeed, the input and solution spaces of face alignment can be effectively divided given auxiliary face attributes. In a small experiment, we average a set of face images according to different attributes, as shown in Fig.~\ref{fig:highlightb}), where the frontal and smiling faces show the mouth corners, while there are no specific details for the image averaged over the whole dataset.
%
%
Given the rich auxiliary information, treating facial landmark detection in isolation is counterproductive.

This study aims to investigate the possibility of optimizing facial landmark detection (the main task) by leveraging auxiliary information from attribute inference tasks. Potential auxiliary tasks include head pose estimation, gender classification, age estimation~\cite{ChenCVPR2013}, facial expression recognition, or facial attribute inference~\cite{aqatagfa23}.
Given the multiple tasks, deep convolutional network appears to be a viable model choice since it allows for joint features learning and multi-objective inference. Typically, one can formulate a cost function that encompasses all the tasks and use the cost function in the network back-propagation learning.
%
%
We show that this conventional multi-task learning scheme is challenging in our problem. There are several reasons. First, the different tasks of face alignment and attribute inference are inherently different in learning difficulties. For instance, learning to identify ``wearing glasses'' attribute is easier than determining if one is smiling. Second, we rarely have auxiliary tasks with similar number of positive/negative cases. For instance, male/female classification enjoys more balanced samples than facial expression recognition.
As a result, different tasks have different convergence rates. In many cases we observe that the joint learning with a specific auxiliary task improves the convergence of landmark detection at the beginning of the training procedure, but become ineffective when the auxiliary task training encounters local minima or over-fitting. Continuing the training with all tasks jeopardizes the network convergence, leading to poor landmark detection performance.

%


%



Our study is the first attempt to demonstrate that face alignment can be jointly optimized with the inference of heterogeneous but subtly correlated auxiliary attributes.
We show that the supervisory signal of auxiliary tasks can be back-propagated jointly with that of face alignment to learn the underlying regularities of face representation.
Nonetheless, the learning is non-trivial due to the different natures and convergence rates of different tasks.
Our key contribution is a newly proposed \textit{Tasks-Constrained Deep Convolutional Network} (TCDCN), with new objective function to address the aforementioned challenges.
In particular, our model considers the following aspects to make the learning effective:
\begin{itemize}
\item \textit{Dynamic task coefficient} -
Unlike existing multi-task deep models~\cite{ahmed2008training,weston2012deep,ABC} that treat all tasks as equally important, we assign and weight each auxiliary task with a coefficient, which is adaptively and dynamically adjusted based on training and validation errors achieved so far in the learning process. Thus a task that is deemed not beneficial to the main task is prevented from contributing to the network learning. This approach can be seen as a principled way of achieving ``early stopping'' on specific task.
In the experiments, we show that the dynamic task coefficient is essential to reach the peak performance for face alignment.
\item \textit{Inter-task correlation modeling} - We additionally model the relatedness of heterogeneous tasks in a covariance matrix in the objective function. Different from the dynamic task coefficient that concerns on the learning convergence, inter-task correlation modeling helps better exploiting the relation between tasks to achieve better feature learning.
\end{itemize}
All the network parameters, including the filters, dynamic task coefficients, and inter-task correlation are learned automatically using a newly proposed alternating optimization approach.

Thanks to the effective shared representation learned from multiple auxiliary attributes, the proposed approach outperforms other deep learning based approaches for face alignment, including the cascaded CNN model~\cite{Sun2013} on five facial point detection.
We demonstrate that shared representation learned by a TCDCN for sparse landmarks can be readily transferred to handle an entirely different configuration with more landmarks, \eg~68 points in the 300-W dataset~\cite{300w}. With the transferred configuration, our method further outperforms other existing methods~\cite{dollar13,Cao2012,6618919,Yu2013,Zhu2012,zhang2014coarse,300w_lbp} on the challenging 300-W dataset, as well as the Helen~\cite{Helen} and COFW~\cite{dollar13} dataset.


In comparison to our earlier version of this work~\cite{zhang2014facial}, we introduce the new dynamic task coefficient to generalize the original idea of task-wise early stopping~\cite{zhang2014facial} (discussed in Sec.~\ref{sec:overview}). Specifically, we show that the dynamic task coefficient is a relatively more effective mechanism to facilitate the convergence of a heterogeneous task network.
In addition, we formulate a new objective function that learns different tasks and their correlation jointly, which further improves the performance and allows us to analyze the usefulness of auxiliary tasks more comprehensively.
Apart from the methodology, the paper was also substantially improved by providing more technical details and more extensive experimental evaluations.

\section{Related Work}
\label{sec:related_work}

\noindent \textbf{Facial landmark detection}:
Conventional facial landmark detection methods can be divided into two categories, namely regression-based method and template fitting method. A regression-based method estimates landmark locations explicitly by regression using image features. For example, Valstar~\etal~\cite{Valstar2010} predict landmark location from local image patch with support vector regression. Cao~\etal~\cite{Cao2012} and Burgos-Artizzu~\etal~\cite{dollar13}  employ cascaded fern regression with pixel-difference features. A number of studies~\cite{dollar2010cascaded,Cootes2012,Dantone2012,yang2013sieving,300w_lbp,chen2014joint} use random regression forest to cast votes for landmark location based on local image patch with Haar-like features. Most of these methods refine an initial guess of the landmark location iteratively, the first guess/initialization is thus critical. By contrast, our deep model takes raw pixels as input without the need of any facial landmark initialization. Importantly, our method differs in that we exploit auxiliary tasks to facilitate landmark detection learning.

A template fitting method builds face templates to fit input images~\cite{Cootes2001,Liu2007,pedersoli2014using}. Part-based model has recently been used for face fitting~\cite{asthana2013robust,Yu2013,Zhu2012}. Zhu and Ramanan~\cite{Zhu2012} show that face detection, facial landmark detection, and pose estimation can be jointly addressed. Our method differs in that we do not limit the learning of specific tasks, \ie~the TCDCN is readily expandable to be trained with additional auxiliary tasks. Specifically, apart from pose, we show that other facial attributes such as gender and expression, can be useful for learning a robust landmark detector. Another difference to~\cite{Zhu2012} is that we learn feature representation from raw pixels rather than pre-defined HOG as face descriptor.

\vspace{0.1cm}
\noindent \textbf{Landmark detection by deep learning}:
The methods~\cite{DeepRegression,zhang2014coarse,Sun2013} that use deep learning for face alignment are close to our approach. The methods usually formulate the face alignment as a regression problem and use multiple deep models to locate the landmarks in a coarse-to-fine manner, such as the cascaded CNN by Sun~\etal~\cite{Sun2013}. The cascaded CNN requires a pre-partition of faces into different parts, each of which are processed by separate deep CNNs. The resulting outputs are subsequently averaged and channeled to separate cascaded layers to process each facial landmark individually. Similarly, Zhang~\etal\cite{zhang2014coarse} uses successive auto-encoder networks to perform coarse-to-fine alignment. Instead, our model requires neither pre-partition of faces nor cascaded networks, leading to drastic reduction in model complexity, whilst still achieving comparable or even better accuracy. This opens up possibility of application in computational constrained scenario, such as the embedded systems. In addition, the use of auxiliary task can reduce the overfitting problem of deep model because the local minimum for different tasks might be in different places. Another important difference is that our method performs feature extraction in the whole face image automatically, instead of handcraft local regions.

\vspace{0.1cm}

\noindent \textbf{Learning multiple tasks in neural network}:
Multitask learning (MTL) is the process of learning several tasks simultaneously with the aim of mutual benefit. This is an old idea in machine learning. Caruana~\cite{caruana1997multitask} provides a good overview focusing on neural network.
Deep model is well suited for learning multiple tasks since it allows for joint features learning and multi-objective inference. Joint learning of multiple tasks has also proven effective in many computer vision problems~\cite{ahmed2008training,weston2012deep,ABC}. However, existing deep models~\cite{collobert2008unified,ahmed2008training} are not suitable to solve our problem because they assume similar learning difficulties and convergence rates across all tasks. For example, in the work of~\cite{ABC}, the algorithm simultaneously learns a human pose regressor and multiple body-part detectors. This algorithm optimizes multiple tasks directly without learning the task correlation. In addition, it uses pre-defined task coefficients in the iterative learning process. Applying this method on our problem leads to difficulty in learning convergence, as shown in Sec.~\ref{sec:experiments}. We mitigate this shortcoming by introducing dynamic task coefficients in the deep model. This new formulation generalizes the idea of early stopping. Early stopping of neural network can date back to the work of Caruana~\cite{caruana1997multitask}, but it is heuristic and  limited to shallow multilayer perceptrons. The scheme is also not scalable for a large quantity of tasks. Different from the work of~\cite{6753948}, which learns the task priority to handle outlier tasks, the dynamic task coefficient in our approach is based on the training and validation error, and aims to coordinate tasks of different convergence rates. We show that dynamic task coefficient is important for joint learning multiple objectives in deep convolutional network.

\section{Landmark Detection with Auxiliary Attributes}
\label{sec: TC-CNN1}
\begin{figure*}
  \centering
  \includegraphics[width=0.9\textwidth]{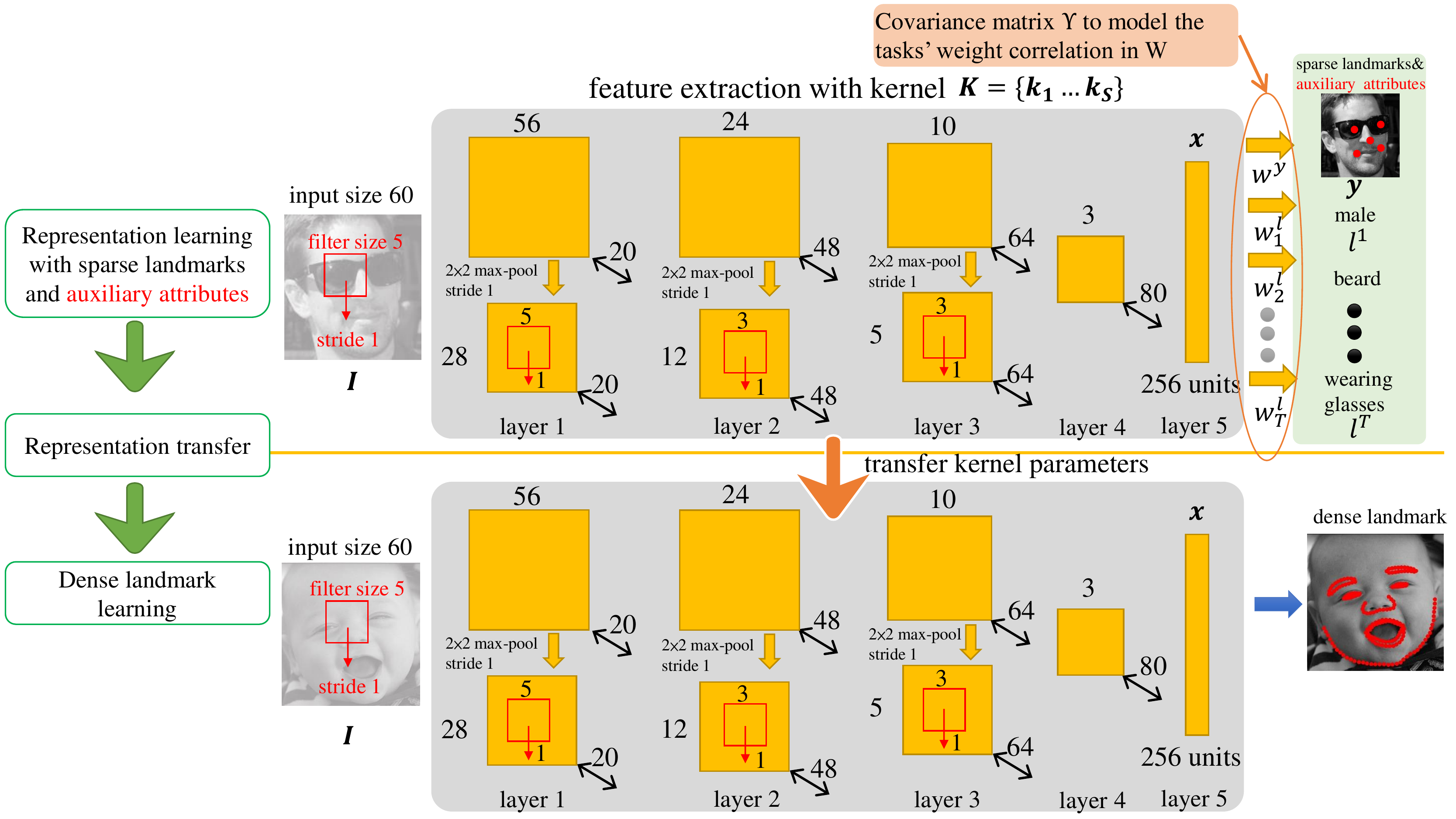}\\
  \caption{Structure specification for TCDCN. A $60\times60$ image is taken as input. In the first layer, we convolve it with 20 different $5\times5$ filters, using a stride of 1. The obtained feature map is $56\times56\times20$, which is subsampled to $28\times28\times20$ with a $2\times2$ max-pooling operation. Similar operations are repeated in layer 2, 3, 4, as the parameters shown in the figure. The last layer is fully-connected. Then the output is obtained by regression.}

  \label{fig:networkstructure}
\end{figure*}

\subsection{Overview}
\label{sec:overview}

We cast facial landmark detection as a nonlinear transformation problem, which transforms the raw pixels of a face image to the positions of dense landmarks.
The proposed framework is illustrated in Fig. \ref{fig:networkstructure},
showing that the highly nonlinear function is modeled as a DCN, which is pre-trained by five landmarks and then fine-tuned to predict the dense landmarks.
Since dense landmarks are expensive to label, the pre-training step is essential because it prevents DCN from over-fitting to small dataset.
In general, the pre-training and fine-tuning procedures are similar, except that the former step initializes filters by a standard normal distribution, while the latter step initializes filters using the pre-trained network.

As shown in Fig. \ref{fig:networkstructure}, DCN extracts a high-level representation $\x\in\mathbb{R}^{D\times1}$ on a face image $I$ using a set of filters $\textbf{K}=\{\kk_s\}_{s=1}^S$, $\x=\phi(I|\textbf{K})$, where $\phi(\cdot)$ is the nonlinear transformation learned by DCN.
%
%
With the extracted feature $\x$, we jointly estimate landmarks and attributes, where landmark detection is the main task and attribute prediction is the auxiliary task.
Let $\{y^m\}_{m=1}^M$ denote a set of real values, representing the x-,y-coordinate of the landmarks, and let $\{l^t\}_{t=1}^T$ denote a set of binary labels of the face attributes, $\forall l^t\in\{0,1\}$.
Specifically, $M$ equals $5\times2=10$ in the pre-training step, implying that the landmarks include two centers of the eyes, nose, and two corners of the mouth.
$M$ represents the number of dense landmarks in the fine-tuning step, such as $M=194\times2=388$ in Helen dataset \cite{Helen} and $M=68\times2=136$ in 300-W dataset \cite{300w}.
This work investigates the effectiveness of 22 attributes in landmark detection, \ie$T=22$.

Both landmark detection and attribute prediction can be learned by the generalized linear models~\cite{mccullagh1989generalized}.
Suppose $\W=[\w^y_1,\w^y_2,...,\w^y_M,\w^l_1,\w^l_2,...,\w^l_T]$ be a weight matrix, $\W\in\mathbb{R}^{D\times{(M+T)}}$, where each column vector corresponds to the parameters of a single task.
For example, $\w^y_2\in\mathbb{R}^{D\times1}$ indicates the parameter vector for the y-coordinate of the first landmark.
With these parameters, we have
\begin{equation}\label{eq:y}
y^m=\trans{\w^y_m}\x+\epsilon_m^y,
\end{equation}
%
where $\epsilon_m^y$ represents an additive random error variable that is distributed according to a normal distribution with mean zero and variance $\sigma_m^2$, \ie $\epsilon_m^y\thicksim\mathcal{N}(0,\sigma_m^2)$.
Similarly, each $\w^l_t$ represents the parameter vector of the $t$-th attribute, which is model as
\begin{equation}\label{eq:l}
l^t=\trans{\w^l_t}\x+\epsilon^l_t,
\end{equation}
where $\epsilon^l_t$ is distributed following a standard logistic distribution, \ie $\epsilon^l_t\thicksim Logistic(0,1)$.



If all the tasks are independent, $\W$ can be simply modeled as a product of the multivariate normal distribution, \ie $\forall\w^y_m,\w^l_t\thicksim\mathcal{N}(\textbf{0},\varepsilon^2\mathcal{I})$, where $\textbf{0}$ is a $D\times1$ zero vector, $\mathcal{I}$ denote a $D\times D$ identity matrix, and $\varepsilon^2$ is a $D\times1$ vector representing the diagonal elements of the covariance matrix.
However, this work needs to explore which auxiliary attribute is crucial to landmark detection, implying that we need to model the correlations between tasks.
Therefore, we assume $\W$ is distributed according to a matrix normal distribution \cite{gupta1999matrix}, \ie $\W\thicksim \mathcal{MN}_{D\times(M+T)}(\mathbf{0},\Upsilon,\varepsilon^2\mathcal{I})$, where $\textbf{0}$ is a $D\times(M+T)$ zero matrix and $\Upsilon$ is a $(M+T)\times(M+T)$ task covariance matrix. The matrix $\Upsilon$ is learned in the training process and can naturally capture the correlation between the weight of different tasks.


As landmark detection and attribute prediction are heterogenous tasks, different auxiliary attribute behaves differently in the training procedure. They may improve the convergence of landmark detection at the beginning of the training procedure, but may become ineffective as training proceeds when local minima or over-fitting is presented.
Thus, each auxiliary attribute is assigned with a dynamic task coefficient $\lambda_t$, $t=1...T$, which is adjusted adaptively during training.
$\lambda_t$ is distributed according to a normal distribution with mean $\mu_t$ and variance $\sigma^2_t$, \ie $\lambda_t\thicksim\mathcal{N}(\mu_t,\sigma^2_t)$, where we assume $\sigma^2_t=1$ and $\mu_t$ is determined based on the training and validation errors (detailed in Sec.~\ref{sec:learningnetwork}).


It is worth pointing out that in in the early version of this work~\cite{zhang2014facial}, we introduce a task-wise early stopping scheme to halt a task after it is no longer beneficial to the main task. This method is heuristic and the criterion to determine when to stop learning a task is empirical. In addition, once a task is halted, it will never resume during the training process.
In contrast to this earlier proposal, the dynamic task coefficient is dynamically updated. Thus a halted task may be resumed automatically if it is found useful again during the learning process. In particular, the dynamic task coefficient has no single optimal solution across the whole learning process. Instead, its value is updated to fit the current training status.



In summary, given a set of face images and their labels, we jointly estimate the filters $\K$, the weight matrix $\W$, the task covariance matrix $\Upsilon$, and the dynamic coefficients $\Lambda=\{\lambda_t\}_{t=1}^T$.

\subsection{Problem Formulation}
\label{sec:problemformulation}

The above problem can be formulated as a probabilistic framework.
Given a data set with $N$ training samples, denoted as $\{\I,\Y,\LL\}$, where $\I=\{I_i\}_{i=1}^N$, $\Y=\{\{y^m_i\}_{m=1}^M\}_{i=1}^N$, and $\LL=\{\{l^t_i\}_{t=1}^T\}_{i=1}^N$, and a set of parameters $\{\K,\W,\Upsilon,\Lambda\}$, we optimize the parameters by maximizing a posteriori probability (MAP)
\begin{equation}\label{eq:map1}
\K^\ast,\W^\ast,\Upsilon^\ast,\Lambda^\ast=\argmax_{\K,\W,\Upsilon,\Lambda} p(\K,\W,\Upsilon,\Lambda|\I,\Y,\LL).
\end{equation}
Eqn.(\ref{eq:map1}) is proportional to
\begin{equation}\label{eq:map2}
\begin{split}
p(\K,\W,\Upsilon,\Lambda|\I,\Y,\LL)\propto~&p(\Y|\I,\K,\W_{M})p(\LL|\I,\K,\W_{T},\Lambda)\cdot\\
&p(\W|\Upsilon)p(\Lambda)p(\K),
\end{split}
\end{equation}
where the first two terms are the likelihood probabilities and the last three terms are the prior probabilities. Moreover, $\W_{M}$ and $\W_{T}$ represent the first $M$ columns and the last $T$ columns of $\W$, respectively. In the following, we will introduce each term of Eqn.(\ref{eq:map2}) in detail.

$\bullet$ The likelihood probability $p(\Y|\I,\K,\W_{M})$ measures the accuracy of landmark detection. As discussed in Eqn.(\ref{eq:y}), each variable of landmark position can be modeled as a linear regression plus a Gaussian noise.
The likelihood can be factorized as
\begin{equation}\label{eq:Y}
p(\Y|\I,\K,\W_{M})=\prod_{i=1}^N\prod_{m=1}^M\mathcal{N}(\trans{\w^y_m}\x_i,\sigma^2_m).
\end{equation}

$\bullet$ The likelihood probability $p(\LL|\I,\K,\W_{T},\Lambda)$ measures the accuracy of attribute prediction. As introduced in Eqn.(\ref{eq:l}), each binary attribute is predicted by a linear function plus a logistic distributed random noise, implying that the probability of $l^t_i$ is a sigmoid function, which is $p(l^t_i=1|\x_i)=f(\trans{\w_t^l}\x_i)$, where $f(x)=1/(1+\exp\{-x\})$.
Thus, the likelihood can be defined as product of Bernoulli distributions
\begin{equation}\label{eq:L}
\begin{split}
p(\LL|\I,\K,\W_{T},\Lambda)&\propto\\\prod_{i=1}^N\prod_{t=1}^T
\{p(l^t_i=1&|\x_i)^{l^t_i}\big(1-p(l^t_i=1|\x_i)\big)^{1-l^t_i}\}^{\lambda_t}.
\end{split}
\end{equation}

$\bullet$ The prior probability of the weight matrix, $p(\W|\Upsilon)$, is modeled by a matrix normal distribution with mean zero \cite{gupta1999matrix}, which is able to capture the correlations between landmark detection and auxiliary attributes.
It is written as
\begin{equation}\label{eq:W}
p(\W|\Upsilon)=\frac{\exp\big\{-\frac{1}{2}\mathrm{tr}[(\varepsilon^2\mathcal{I})^{-1}
\W\Upsilon^{-1}\trans{\W}]\big\}}{(2\pi)^{\frac{D(M+T)}{2}}
|\varepsilon^2\mathcal{I}|^{\frac{M+T}{2}}|\Upsilon|^{\frac{D}{2}}},
\end{equation}
where $\mathrm{tr}(\cdot)$ calculates the trace of a matrix and $\Upsilon$ is a positive semi-definite matrix modeling the task covariance, denoted as $\Upsilon\succeq\textbf{0}$, $\Upsilon\in\mathbb{R}^{(M+T)\times(M+T)}$.
Referring to Eqn.(\ref{eq:W}), the variance between the $m$-th landmark and the $t$-th attribute is obtained by $\sum_{d=1}^D\W_{(d,m)}\Upsilon^{-1}_{(m,m+t)}\W_{(d,m+t)}$, where $\W_{(d,m)}$ denotes the element in the $d$-th row and $m$-th column, showing that the relation of a pair of tasks is measured by their corresponding weights with respect to each feature dimension $d$.
For instance, if two different tasks select or reject the same set of features, they are highly correlated.
More clearly, Eqn.(\ref{eq:W}) is a matrix form of the multivariate normal distribution. They are equivalent if $\W$ is reshaped as a long vector.

$\bullet$ The prior probability of the tasks' dynamic coefficients is defined as a product of the normal
distributions, $p(\Lambda)=\prod_{t=1}^T \mathcal{N}(\mu_t,\sigma_t^2)$, where the mean is adjustable based on the training and validation errors. It has significant difference with the task covariance matrix.
For example, the auxiliary attribute `wearing glasses' is probably related to the landmark positions of eyes. Their relation can be measured by $\Upsilon$. However, if `wearing glasses' converges more quickly than the other tasks, it becomes ineffective because of local minima or over-fitting. Therefore, its dynamic coefficient could be decreased to avoid these side-effects.

$\bullet$ The DCN filters can be initialized as a standard multivariate normal distribution as previous methods \cite{krizhevsky2012imagenet} did. In particular, we define $p(\K)=\prod_{s=1}^Sp(\kk_s)=\prod_{s=1}^S\mathcal{N}(\textbf{0},\mathcal{I})$.

By taking the negative logarithm of Eqn.(\ref{eq:map2}) and combining Eqn.(\ref{eq:Y}), (\ref{eq:L}), and (\ref{eq:W}), we obtain the MAP objective function
\begin{equation}\label{eq:MAP}
\begin{split}
&\argmin_{\K,\W,\Lambda,\Upsilon\succeq\textbf{0}}\sum_{i=1}^N\sum_{m=1}^M(y_i^m-\trans{\w^y_m}\x_i)^2\\
&-\sum_{i=1}^N\sum_{t=1}^T\lambda_t\Big\{ l^t_i\ln f(\trans{\w^l_t}\x_i)+(1-l^t_i)\ln\big(1-f(\trans{\w^l_t}\x_i)\big)\Big\}
\\
&+\mathrm{tr}(\W\Upsilon^{-1}\trans{\W})+D\ln|\Upsilon|
+\sum_{s=1}^S\trans{\kk}_s\kk_s+\sum_{t=1}^T(\lambda_t-\mu_t)^2.
\end{split}
\end{equation}
Eqn.(\ref{eq:MAP}) contains six terms. For simplicity of discussion, we remove the terms that are constant. We also assume the variance parameters such as $\sigma_m, \forall m=1...M$, $\sigma_t,\forall t=1...T$, and $\varepsilon$ equal one. Thus, the regularization parameters of the above terms are comparable and can be simply ignored.

Eqn.(\ref{eq:MAP}) can be minimized by updating one parameter with the remaining parameters fixed.
First, although the first three terms are likely to be jointly convex with respect to $\W$, $\x_i$ in the first two terms is a highly nonlinear transformation with respect to $\K$, \ie $\x_i=\Phi(I_i|\K)$.
In this case, no global optima are guaranteed. Therefore, following the optimization strategies of CNN \cite{lecun1998gradient}, we apply stochastic gradient descent (SGD) \cite{krizhevsky2012imagenet} with weight decay \cite{moody1995simple} to search the suitable local optima for both $\W$ and $\K$.
This method has been demonstrated working reasonably well in practice \cite{krizhevsky2012imagenet}.
Here, the fifth term can be considered as the weight decay of the filters.
Second, the third term in Eqn.(\ref{eq:MAP}) is a convex function regarding $\Upsilon$, but the fourth term is concave since negative logarithm is a convex function.
In other words, learning $\Upsilon$ directly is a convex-concave problem \cite{yuille2002concave}.
However, with a well-known lemma \cite{boyd2004convex}, $\ln|\Upsilon|$ has a convex upper bound, $\ln|\Upsilon|\leq\mathrm{tr}(\Upsilon)-M-T$. Thus, the fourth term can be replaced by $D\mathrm{tr}(\Upsilon)$. Both the third and the fourth terms are now convex regarding $\Upsilon$.
Finally, since the dynamic coefficients in Eqn.(\ref{eq:MAP}) are linear and independent, finding each $\lambda_t$ has a closed form solution.

\begin{figure*}[t]
  \centering
  \includegraphics[width=0.82\textwidth]{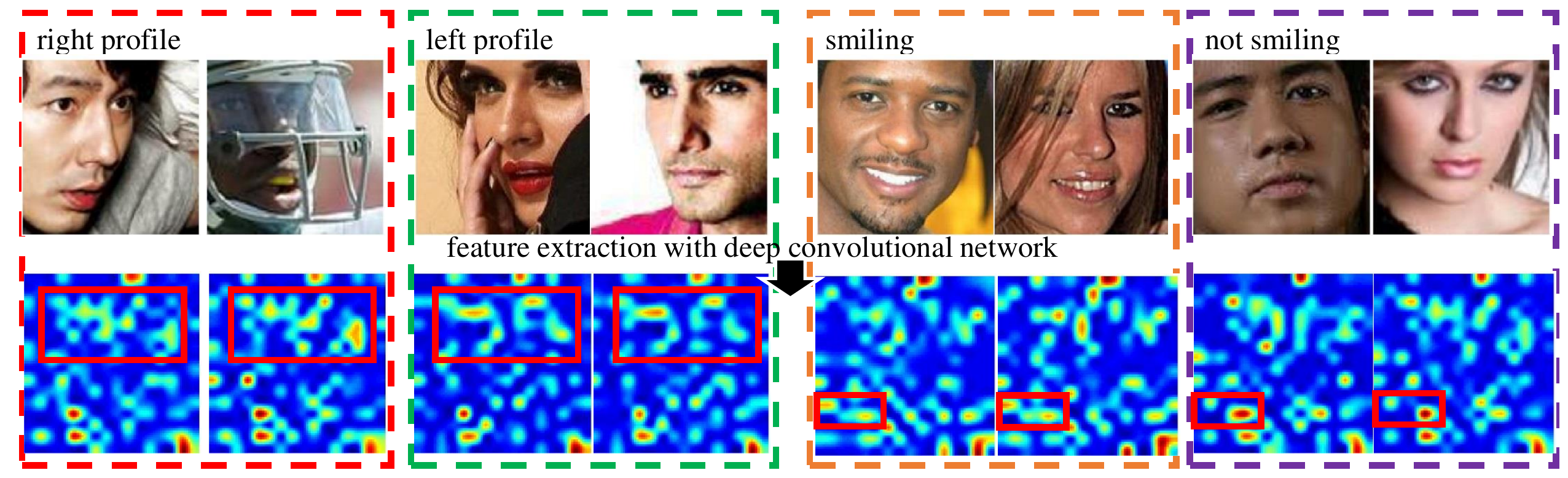}\\
  \vskip -0.3cm
  \caption{The TCDCN learns shared features for facial landmark detection and auxiliary tasks. The first row shows the face images and the second row shows the corresponding features in the shared feature space, where the face images with similar poses and attributes are close with each other. This reveals that the learned feature space is robust to pose, expression, and occlusion.
}\label{fig:features}
\end{figure*}
\subsection{Learning Algorithm}
\label{sec:learningnetwork}

We solve the MAP problem in an iterative manner. First, we jointly update the DCN filters $\K$ and the weight matrix $\W$ with the tasks' dynamic coefficients $\Lambda$ and covariance matrix $\Upsilon$ fixed.
Second, we update the covariance matrix $\Upsilon$ by fixing all the other parameters with their current values.
Third, we update $\Lambda$ in a similar way to the second step.

\vspace{0.1cm}
\textbf{In the first step}, we optimize $\W$ and $\K$ in the DCN and fix $\Upsilon$ and $\Lambda$ with their current values. In this case, the fourth and the last terms in Eqn.(\ref{eq:MAP}) are constant and thus can be removed. We write the loss function in a matrix form as follows
\begin{equation}\label{eq:X}
\begin{split}
E(\I)=\sum_{i=1}^N&\Big\{||\y_i-\trans{\W}_{M}\x_i||^2-
\trans{\bm{\ell}}_i\mathrm{diag}(\Lambda)\ln f(\trans{\W}_{T}\x_i)\\
&-\trans{(\textbf{1}-\bm{\ell}_i)}\mathrm{diag}(\Lambda)\ln\big(\textbf{1}-f(\trans{\W}_{T}\x_i)\big)\Big\}
\\&+\mathrm{tr}(\W\Upsilon^{-1}\trans{\W})+\mathrm{tr}(\K\trans{\K}),
\end{split}
\end{equation}
where $\y$ is a $M\times1$ vector, and $\bm{\ell}$, $\textbf{1}$ are both $T\times1$ vectors. $\mathrm{diag}(\Lambda)$ represents a diagonal matrix with $\lambda_1,...,\lambda_T$ being the values in the diagonal.
The fourth term in Eqn.(\ref{eq:X}) can be considered as the parameterized weight decay of $\W$, while the last term is the weight decay of the filters $\K$, \ie $\mathrm{tr}(\K\trans{\K})=\sum_{s=1}^S\trans{\kk}_s\kk_s$.
Eqn.(\ref{eq:X}) combines the least square loss and the cross-entropy loss to learn the DCN, which can be optimized by SGD \cite{krizhevsky2012imagenet}, since they are defined over individual sample.
Fig. \ref{fig:networkstructure} illustrates the architecture of DCN, containing four convolutional layers and one fully-connected layer. This architecture is a tradeoff between accuracy of landmark detection and computational cost, and it works well in practice. Note that the learning method introduced in this work is naturally compatible with any deep network structure, but exploring them is out of the scope of this paper.

Now we introduce the learning procedure. At the very beginning, each column of $\W$ and each filter of $\K$ are initialized according to a multivariate standard normal distribution. To learn the weight matrix $\W$, we calculate its derivative, $\Delta\W=-\eta\frac{\partial E}{\partial\W}=-\eta\frac{\partial E}{\partial \mathbf{o}}\frac{\partial \mathbf{o}}{\partial f}\frac{\partial f}{\partial\W}$, where $\mathbf{o}$ and $\eta$ denote the network outputs (predictions) and the step size of the gradient descent, respectively. By simple derivation, we have
\begin{eqnarray}
\frac{\partial E}{\partial\W_M}&=&\x_i\trans{(\y_i-\mathbf{o}_i)},\label{eq:dewm}\\
\frac{\partial E}{\partial\W_T}&=&\x_i\trans{(\bm{\ell}_i-\mathbf{o}_i)}\mathrm{diag}(\Lambda),\label{eq:dewt}
\end{eqnarray}
where $\mathbf{o}_i$ is the corresponding tasks' predictions. For example, $\mathbf{o}_i=\trans{\W}_M\x_i$ in Eqn.(\ref{eq:dewm}) indicates the predictions of the landmark positions, while $\mathbf{o}_i=f(\trans{\W}_T\x_i)$ in Eqn.(\ref{eq:dewt}) indicates the predictions of auxiliary attributes. In summary, the entire weight matrix in the $(j+1)$-th iteration is updated by $\W_{j+1}=\W_j-\eta_1\frac{\partial E}{\partial\W_j}-\eta_2(2\W_j\Upsilon^{-1})$, where $\eta_1,\eta_2$ are the regularization parameters of the gradient and the weight decay.

To update filters $\K$, we propagate the errors of DCN from top to bottom, following the well-known back-propagation (BP) strategy \cite{Rumelhart:1988}, where the gradient of each filter is computed by the cross-correlation between the corresponding input channel and the error map \cite{lecun1998gradient}. In particular, at the fully-connected layer as shown in Fig. \ref{fig:networkstructure}, the errors are obtained by first summing over the losses of both landmark detection and attribute predictions, and then the sum is multiplied by the transpose of the weight matrix. For each convolutional layer, the errors are achieved by the de-convolution \cite{lecun1998gradient} between its filters and the back-propagated errors. Several pairs of face images and their features obtained by filters $\K$ are shown in Fig.~\ref{fig:features}, which shows that the learned features are robust to large poses and expressions. For example, the features of smiling faces or faces have similar poses exhibit similar patterns.

\vspace{0.1cm}
\textbf{In the second step}, we optimize the covariance matrix $\Upsilon$ with $\W$, $\K$, and $\Lambda$ fixed. As discussed in Eqn.(\ref{eq:MAP}), the logarithm of $\Upsilon$ can be relaxed by its upper bound. The optimization problem for finding $\Upsilon$ then becomes
\begin{equation}\label{eq:Upsilon}
\begin{split}
&\min_{\Upsilon} \,\, \textrm{tr}(\W\Upsilon^{-1}\trans{\W})\\
&\textrm{s.t.}\,\,\,\,\,\,\Upsilon\succeq\textbf{0}, \,\,\,\textrm{tr}(\Upsilon)\leq\eta.\\
\end{split}
\end{equation}
For simplicity, we assume $\eta=1$. Problem (\ref{eq:Upsilon}) with respect to $\Upsilon$ is a naive semi-definite programming problem and has a simple closed form solution, which is $\Upsilon=\frac{(\trans{\W}\W)^{\frac{1}{2}}}
{\mathrm{tr}\big((\trans{\W}\W)^{\frac{1}{2}}\big)}$.


\vspace{0.1cm}
\textbf{In the third step}, we update the dynamic coefficients $\Lambda$ with $\W$, $\K$, and $\Upsilon$ fixed. By ignoring the constant terms in Eqn.(\ref{eq:MAP}), the optimization problem becomes
\begin{equation}\label{eq:Lambda}
\begin{split}
\min_{\Lambda}&\frac{1}{N}\sum_{i=1}^N\sum_{t=1}^T-\lambda_t\{ l^t_i\ln f(\trans{\w^l_t}\x_i)\\
&+(1-l^t_i)\ln\big(1-f(\trans{\w^l_t}\x_i)\big)\}+\frac{1}{2}\sum_{t=1}^T(\lambda_t-\mu_t)^2,\\
&s.t.~~~1\geq\lambda_t\geq\epsilon,~t=1,2,...,T
\end{split}
\end{equation}
where $\epsilon$ is a small constant close to zero. Each $\lambda_t$ has a analytical solution, which is $\lambda_t=\min\big\{1,\max\big\{\epsilon,~\mu_t+\frac{1}{N}\sum_{i=1}^Nl^t_i\ln f(\trans{\w^l_t}\x_i)+(1-l^t_i)\ln\big(1-f(\trans{\w^l_t}\x_i)\big)\big\}\big\}$, implying that each dynamic coefficient is determined by its expected value and the loss value averaged over $N$ training samples. Here, we can define $\mu_t$ similar to the task-wise early stopping \cite{zhang2014facial}. Suppose the current iteration is $j$, let $E_{val}^t(j)$, and $E_{tr}^t(j)$ be the values of the loss function of task $t$ on the validation set and training set, respectively. We can have
\begin{equation}\label{eq:criteria}
\begin{split}
\mathbf{\mu}_{t} = \rho\times\frac{E^{t}_{val}(j-\tau)-E^{t}_{val}(j)}{E^{t}_{val}(j-\tau)}\times
\frac{E^{t}_{tr}(j-\tau)-E^{t}_{tr}(j)}{E^{t}_{tr}(j-\tau)},
\end{split}
\end{equation}
where $\rho$ is a constant scale factor, and $\tau$ controls a training strip of length $\tau$. The second term in Eqn.(\ref{eq:criteria}) represents the tendency of the validation error. If the validation error drops rapidly within a period of length $\tau$, the value of the first term is large, indicating that training should be emphasized as the task is valuable. Similarly, the third term measures the tendency of the training error. We can see that the task-wise early stopping strategy proposed in\cite{zhang2014facial} can be treated as a special case of the dynamic coefficient $\mathbf{\lambda}_{t}$. In addition, we does not need a tuned threshold to decide whether to stop a task as\cite{zhang2014facial}, and we can provide better performance (see Sec.~\ref{sec:exp_dynamic_task}).

\subsection{Transferring TCDCN for Dense Landmarks}
\label{sec:300w}
 After training the TCDCN model on sparse landmarks and auxiliary attributes, it can be readily transferred from sparse landmark detection to handle more landmark points, \eg~68 points as in 300-W dataset~\cite{300w}. In particular, we initialize the network (\ie, the lower part of Fig.~\ref{fig:networkstructure}) with the learned shared representation and fine-tune using a separate training set only labeled with dense landmark points. Since the shared representation of the pre-trained TCDCN model already captures the information from attributes, the auxiliary tasks learning is not necessary in the fine-tuning stage.

\section{Implementation and Experiments}

\label{sec:experiments}
\textbf{Network Structure.} Fig.~\ref{fig:networkstructure} shows the network structure of TCDCN. The input of the network is $60\times60$ gray-scale face image (normalized to zero-mean and unit-variance). The feature extraction stage contains four convolutional layers, three pooling layers, and one fully connected layer. The kernels in each convolutional layer produce multiple feature maps. The commonly used rectified linear unit is selected as the activation function. For the pooling layers, we conduct max-pooling on non-overlap regions of the feature map. The fully connected layer following the fourth convolutional layer produces a feature vector that is shared by the multiple tasks in the estimation stage.

\noindent\textbf{Evaluation metrics}:
In all cases, we report our results on two popular metrics~\cite{dollar13,Cao2012,Dantone2012,Sun2013}, \ie mean error and failure rate. The mean error is measured by the distances between estimated landmarks and the ground truths, and normalized with respect to the inter-ocular distance. Mean error larger than 10\% is reported as a failure.

\subsection{Datasets}
\label{sec:datasets}
\begin{figure*}[t]
  \centering
  \includegraphics[width=0.8\textwidth]{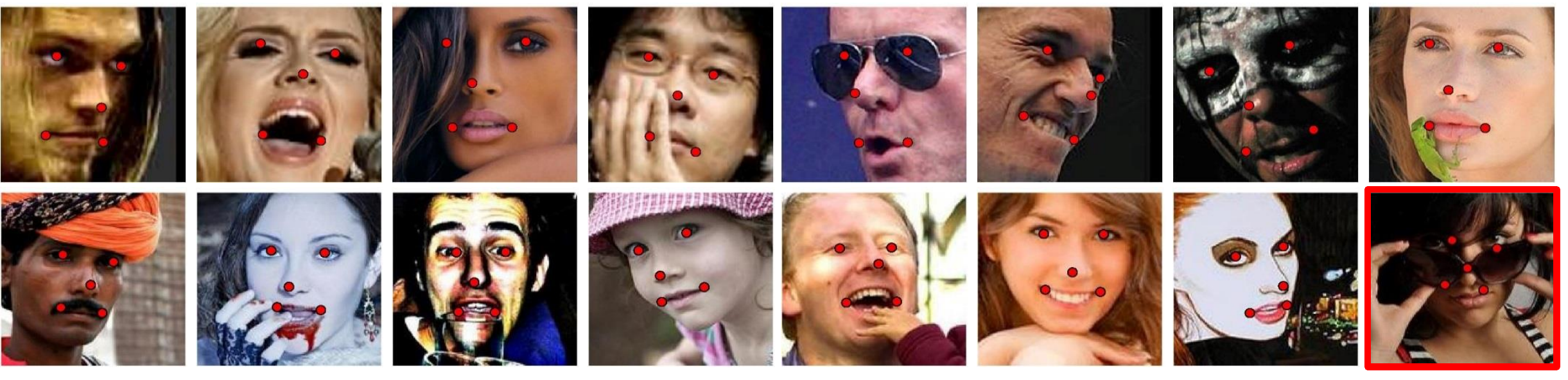}
  \vskip -0.3cm
  \caption{Example alignment results on MAFL dataset (the first row), and AFLW~\cite{Kostinger2011} datasets (the second row). Red rectangles indicate wrong cases.}
  \label{fig:visual_examples_MTFL_AFLW}
  \vskip -0.3cm
\end{figure*}

\vspace{0.1cm}
\noindent\textbf{Multi-Attribute Facial Landmark (MAFL)}\footnote{Data and codes of this work are available at \url{http://mmlab.ie.cuhk.edu.hk/projects/TCDCN.html}. (Ping Luo is the corresponding author)}: To facilitate the training of TCDCN, we construct a new dataset by annotating 22 facial attributes on 20,000 faces randomly chosen from the Celebrity face dataset~\cite{sun2014deep}. The attributes are listed in Table~\ref{tab:attributes} and all the attributes are binary, indicating the attribute is presented or not. We divide the attributes into four groups to facilitate the following analyses. The grouping criterion is based on the main face region influenced by the associated attributes. In addition, we divide the face into one of five categories according to the degree of yaw rotation. This results in the fifth group named as ``head pose''.
%
%
All the faces in the dataset are accompanied with five facial landmarks locations (eyes, nose, and mouth corners), which are used as the target of the face alignment task. We randomly select 1,000 faces for testing and the rest for training. Example images are provided in Fig.~\ref{fig:visual_examples_MTFL_AFLW}.

\vspace{0.1cm}
\noindent\textbf{Annotated Facial Landmarks in the Wild (AFLW)~\cite{Kostinger2011}:} AFLW contains 24,386 face images gathered from Flickr. This dataset is selected because it is more challenging than other conventional datasets, such as BioID~\cite{Jesorsky} and LFPW~\cite{Belhumeur2011}. Specifically, AFLW has larger pose variations (39\% of faces are non-frontal in our testing images) and severe partial occlusions. Each face is annotated with 21 landmarks at most. Some landmarks are not annotated due to out-of-plane rotation or occlusion. We randomly select 3,000 faces for testing. Fig.~\ref{fig:visual_examples_MTFL_AFLW} depicts some examples.

\vspace{0.1cm}
\noindent\textbf{Caltech Occluded Faces in the Wild (COFW)~\cite{dollar13}:} This dataset is collected from the web. It is designed to present faces in occlusions due to pose, the use of accessories (\eg, sunglasses), and interaction with objects (\eg, food, hands). This dataset includes 1,007 faces, annotated with 29 landmarks, as shown in Fig.~\ref{fig:visual_examples_COFW}.

\vspace{0.1cm}
\noindent\textbf{Helen~\cite{Helen}:} Helen contains 2,330 faces from the web, annotated densely with 194 landmarks (Fig.~\ref{fig:visual_examples_helen}).

\vspace{0.1cm}
\noindent\textbf{300-W~\cite{300w}:} This dataset is well-known as a standard benchmark for face alignment. It is a collection of 3,837 faces from existing datasets: LFPW~\cite{Belhumeur2011}, AFW~\cite{Zhu2012}, Helen~\cite{Helen} and XM2VTS~\cite{XM2vts}. It also contains faces from an additional subset called IBUG, consisting images with difficult poses and expressions for face alignment, as shown in Fig.~\ref{fig:visual_examples_300W}. Each face is densely annotated with 68 landmarks.

\begin{table}[t]
\newcommand{\tabincell}[2]{\begin{tabular}{@{}#1@{}}#2\end{tabular}}
\caption{Annotated Face Attributes in MAFL Dataset}
\vskip -0.5cm
\label{tab:attributes}
\begin{center}
\begin{tabular}{l|c}
\hline
Group&Attributes\\
\hline
\hline
eyes&\tabincell{c}{bushy eyebrows, arched eyebrows, narrow eyes,\\ bags under eyes, eyeglasses}\\
\hline
nose&\tabincell{c}{big nose, pointy nose}\\
\hline
mouth&\tabincell{c}{mouth slightly open, no beard, smiling, \\big lips, mustache}\\
\hline
global&\tabincell{c}{gender, oval face, attractive, \\heavy makeup, chubby}\\
\hline
head pose&\tabincell{c}{frontal, left, left profile, right, right profile}\\
\hline
\end{tabular}
\end{center}
\vskip -0.5cm
\end{table}

\subsection{Training with Dynamic Task Coefficient}
\label{sec:exp_dynamic_task}
\begin{figure}[t]
  \centering
  \includegraphics[width=0.5\textwidth]{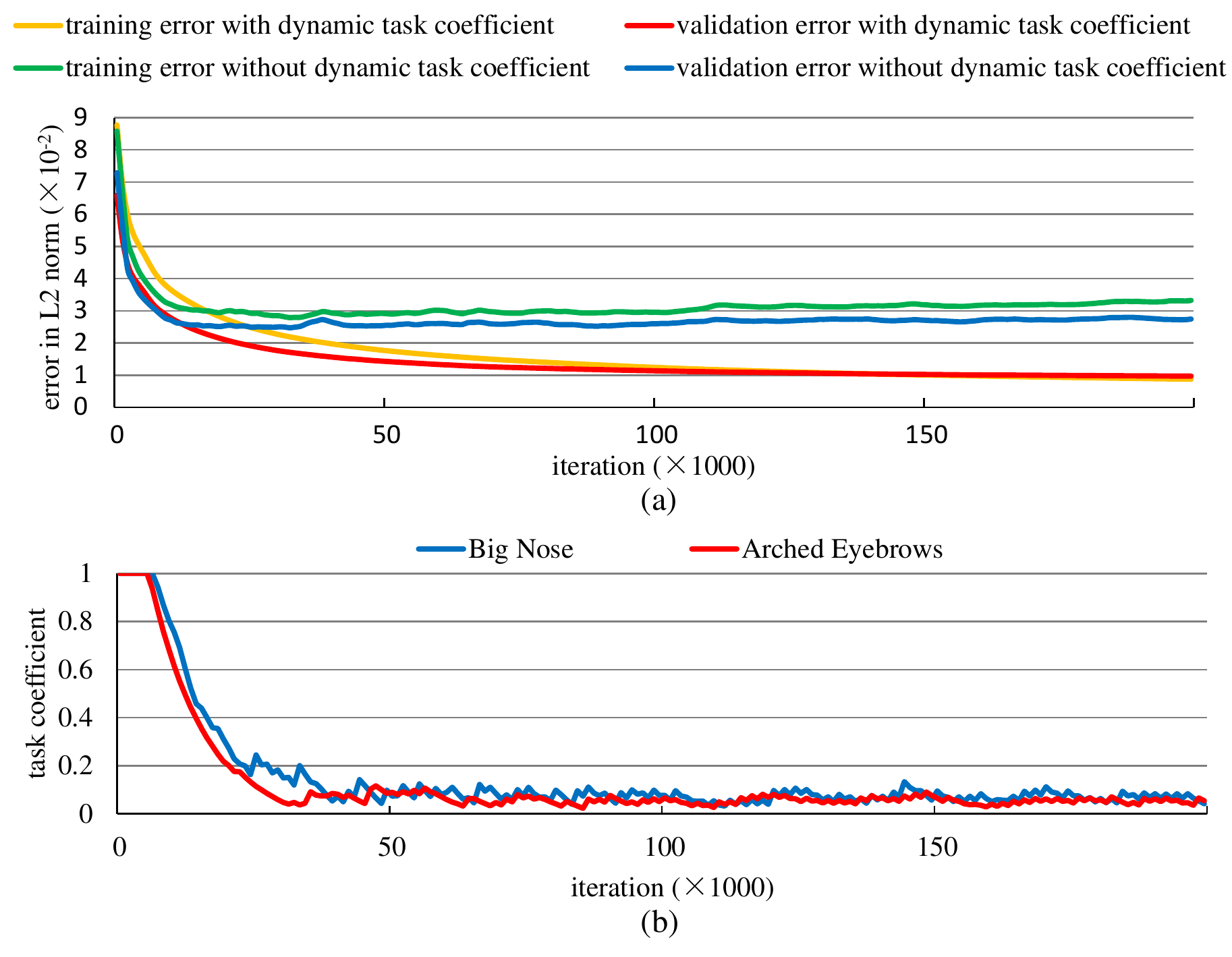}
  \vskip -0.3cm
  \caption{(a) Facial landmark localization error curve with and without dynamic task coefficient.  The error is measured in L2-norm with respect to the ground truth of the 10 coordinates values (normalized to [0,1]) for the 5 landmarks. (b) Task coefficients for the ``Big Nose'' and ``Arched Eyebrows'' attributes over the training process.}
  \label{fig:training_curve}
  \vskip -0.25cm
\end{figure}

Dynamic task coefficient is essential in TCDCN to coordinate the learning of different tasks with different convergence rates.
To verify its effectiveness, we train the proposed TCDCN with and without this technique. Fig.~\ref{fig:training_curve} (a) plots the main task's error of the training and validation sets up to 200,000 iterations. Without dynamic task coefficient, the training error converges slowly and exhibits substantial oscillations. In contrast, convergence rates of both the training and validation sets are fast and stable when using the proposed dynamic task coefficient. In addition, we illustrate the dynamic task coefficients of two attributes in Fig.~\ref{fig:training_curve} (b). We observe that the values of their coefficients drop after a few thousand iterations, preventing these auxiliary tasks from over-fitting. The coefficients may increase when these tasks become effective in the learning process, as shown by the sawtooth-like pattern of the coefficient curves. These two behaviours work together, facilitating the smooth convergence of the main task, as shown in Fig.~\ref{fig:training_curve} (a).

In addition, we compare the dynamic tasks coefficient with the task-wise early stopping proposed in the earlier version of this work~\cite{zhang2014facial}.
As shown in Table~\ref{tab:conf}, dynamic task coefficient achieves better performance than the task-wise early stopping scheme. This is because the new method is more dynamic in coordinating the different auxiliary tasks across the whole training process (see Sec.~\ref{sec:overview}).
%

\begin{table}[t]
\caption{Comparison of mean error ($\times10^{-2}$) on MAFL dataset under different network configurations}
\vskip -0.35cm
\label{tab:conf}
\begin{center}
	\setlength{\tabcolsep}{.0667em}
\begin{tabular}{c|c|c}
\hline
&\parbox{2.5cm}{without inter-task\\correlation learning}&\parbox{2.5cm}{with inter-task\\correlation learning}\\
\hline\hline
task-wise early stopping~\cite{zhang2014facial}&8.35&8.21\\
dynamic task coefficient&8.07&7.95\\
\hline
\end{tabular}
\end{center}
\end{table}

\subsection{Inter-task Correlation Learning}
\begin{figure}
  \centering
  \includegraphics[width=0.45\textwidth]{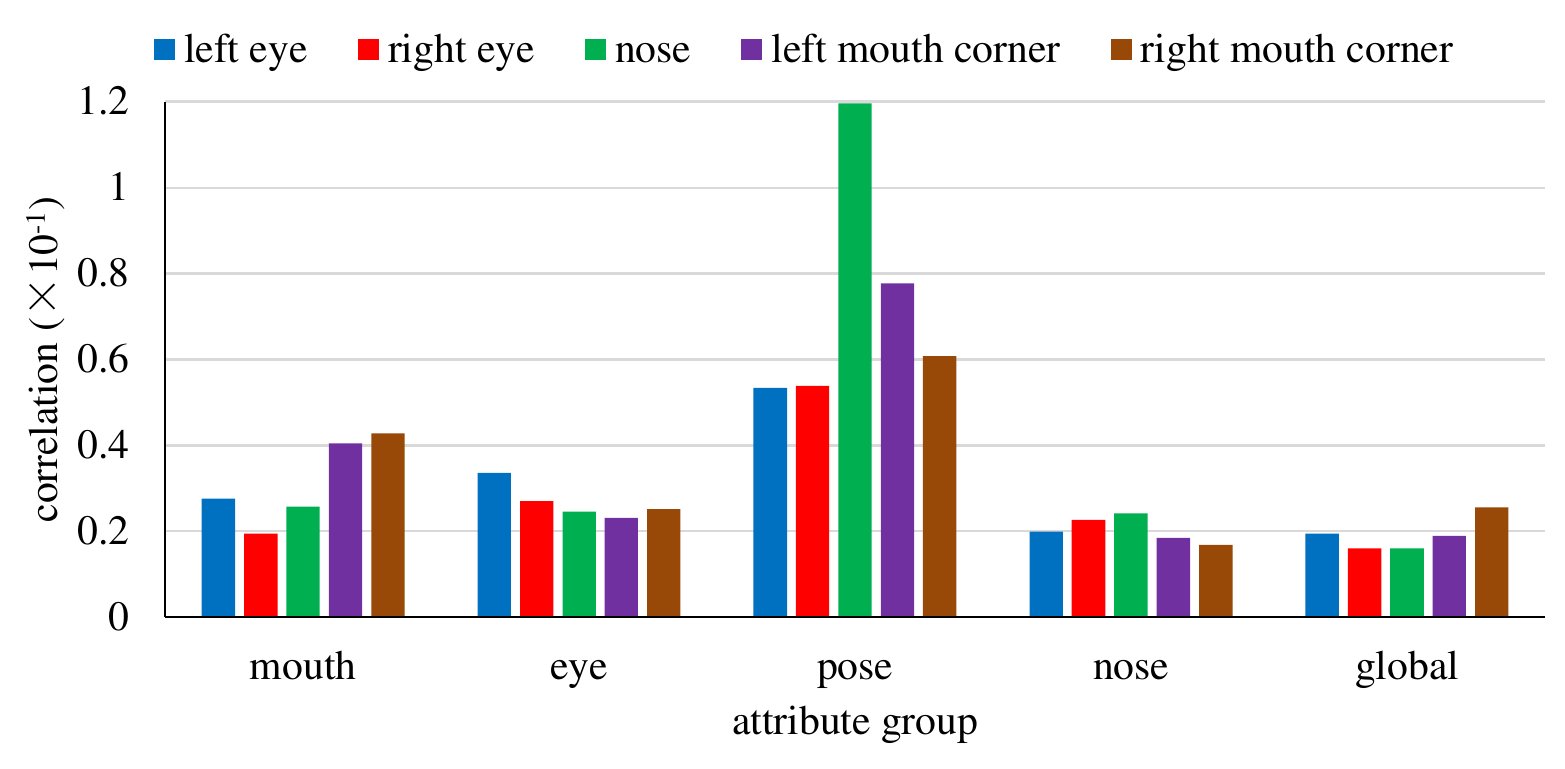}
  \vskip -0.35cm
  \caption{Correlation of each attribute group with different landmarks.}
  \label{fig:attribute_coeff}
\end{figure}

\begin{figure}
  \centering
  \includegraphics[width=0.45\textwidth]{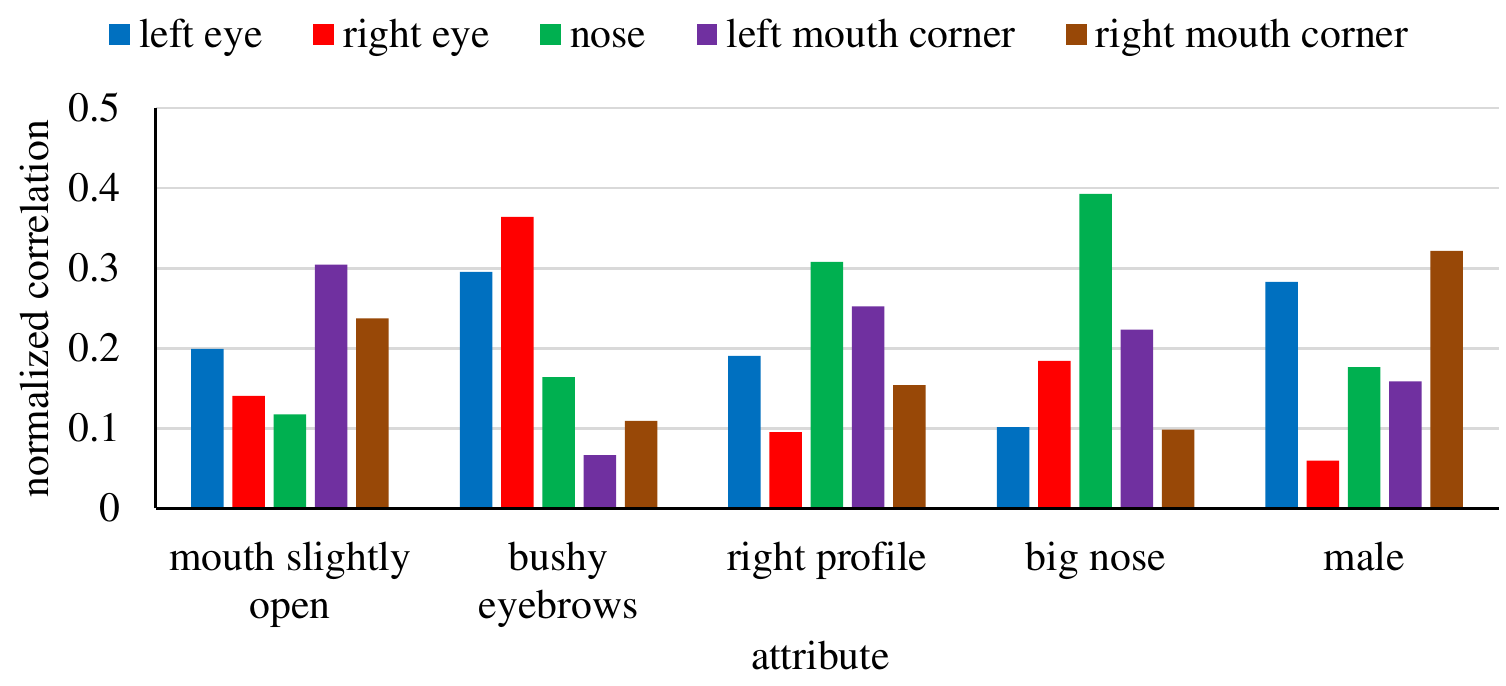}
    \vskip -0.35cm
  \caption{Normalized correlation of the attributes with different landmarks. The attributes are randomly selected from each attribute group. The correlation is normalized among the five landmarks.}
  \label{fig:attribute_coeff2}
  \vskip -0.3cm
\end{figure}

To investigate how the auxiliary tasks help facial landmark detection, we study the learned correlation between these tasks and the facial landmarks. In particular, as we have learned the task covariance matrix $\Upsilon$, given the relation between correlation matrix and covariance matrix, we can compute the correlation between any two tasks, by normalizing their covariance with the square root of the product of their variances. In Fig.~\ref{fig:attribute_coeff}, we present the learned correlation between the attribute groups and facial landmarks. In particular, for each attribute group, we compute the average absolute value of the correlation with the five facial landmarks, respectively. It is shown that for the group of ``mouth'', the correlations with the according landmarks (\ie, mouth corners) are higher than the others. Similar trends can be observed in the group of ``nose'' and ``eyes''. For the group of ``global'', the correlations are roughly even for different landmarks because the attributes are determined by the global face structure. The correlation of the ``pose'' group is much higher than that of the others. This is because the head rotation directly affects the landmark distribution. Moreover, in Fig.~\ref{fig:attribute_coeff2}, we randomly choose one attribute from each attribute group and visualize its correlation to other landmarks. For clarification, for each attribute, we normalize the correlation among the landmarks (\ie, the sum of the correlation on the five landmarks equals one). We can also observe that the attributes are more likely to be correlated to its according landmarks.

\begin{figure}[t]
  \centering
  \includegraphics[width=0.42\textwidth]{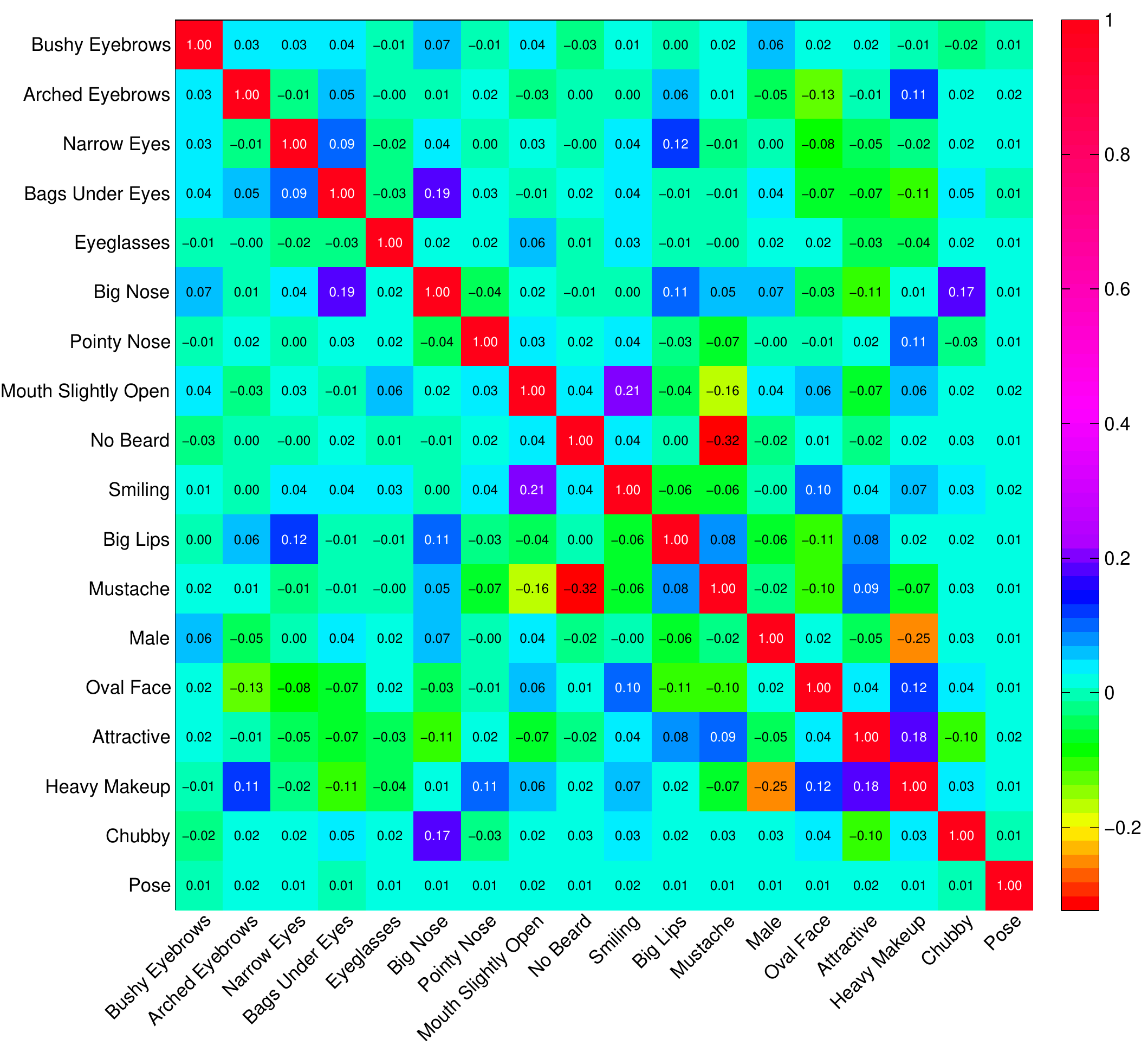}
  \vskip -0.3cm
  \caption{Pairwise correlation of the auxiliary tasks learned by TCDCN (best viewed in color).}
  \label{fig:attr_coeff}
  \vskip -0.3cm
\end{figure}

In addition, we visualize the learned correlation between the auxiliary tasks in Fig.~\ref{fig:attr_coeff}. Because the attributes of ``Left Profile'', ``Left'', ``Frontal'', ``Right'', ``Right Profile'' are mutually exclusive (i.e., only one attribute can be positive for a face) and describe the yaw rotation, we aggregate these five attributes as one attribute (\ie ``pose''), by computing the average absolute correlation with other attributes. One can observe some intuitive results in this figure. For examples, the head pose is unrelated to other attributes; ``Heavy Makeup'' has high positive correlation with ``Attractive'', and high negative correlation with ``Male''. In Table~\ref{tab:conf}, we show the mean errors of facial landmark localization on MAFL dataset with and without inter-task correlation learning (without correlation learning means that we simply apply multiple tasks as targets and do not use the term of $\Upsilon$ in Eq.~(\ref{eq:MAP})). It demonstrates the effectiveness of task correlation learning.

\subsection{Evaluating the Effectiveness of Auxiliary Task}
\label{subset:exp_related_task}
To further examine the influence of auxiliary tasks more comprehensively, we evaluate different variants of the proposed model. In particular, the first variant is trained only on facial landmark detection. We train another five model variants on facial landmark detection along with the auxiliary tasks in the groups of ``eyes'', ``nose'', ``mouth'', ``global'', ``head pose'', respectively. In addition, we synthesize a task with random objective and train it along with the facial landmark detection task, which results in the sixth model variant. The full model is trained using all the attributes. For simplicity, we name each variant by facial landmark detection (FLD) and the auxiliary tasks, such as ``FLD only'', ``FLD+eyes'', ``FLD+pose'', ``FLD+all''.

\begin{figure*}
  \centering
  \includegraphics[width=0.87\textwidth]{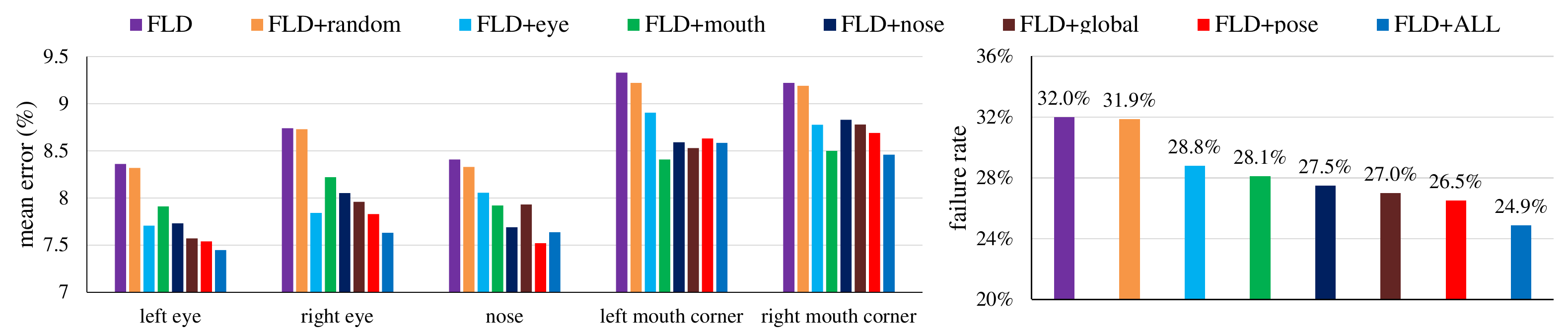}\\
  \vskip -0.3cm
  \caption{Comparison of different model variants of TCDCN: the mean error over different landmarks (left), and the overall failure rate (right).}
  \label{fig:model_variant}
  \vskip -0.5cm
\end{figure*}

It is evident from Fig.~\ref{fig:model_variant} that optimizing landmark detection with auxiliary tasks is beneficial.
In particular, ``FLD+all'' outperforms ``FLD'' by a large margin, with a reduction of over 7\% in failure rate.
When single auxiliary task group is present, ``FLD+pose'' and ``FLD+global'' perform better than the others.
This is not surprising since the pose variation affects locations of all landmarks directly and the ``global' attribute group influences the whole face region.
%
%
%
The other auxiliary tasks such as ``eyes'' and ``mouth'' are observed to have comparatively smaller influence to the final performance, since they mainly capture local information of the face.
As for ``FLD+random'' the performance is hardly improved. 
This result shows that the main task and auxiliary task need to be related for any performance gain in the main task.

\begin{figure}
  \centering
  \includegraphics[width=0.5\textwidth]{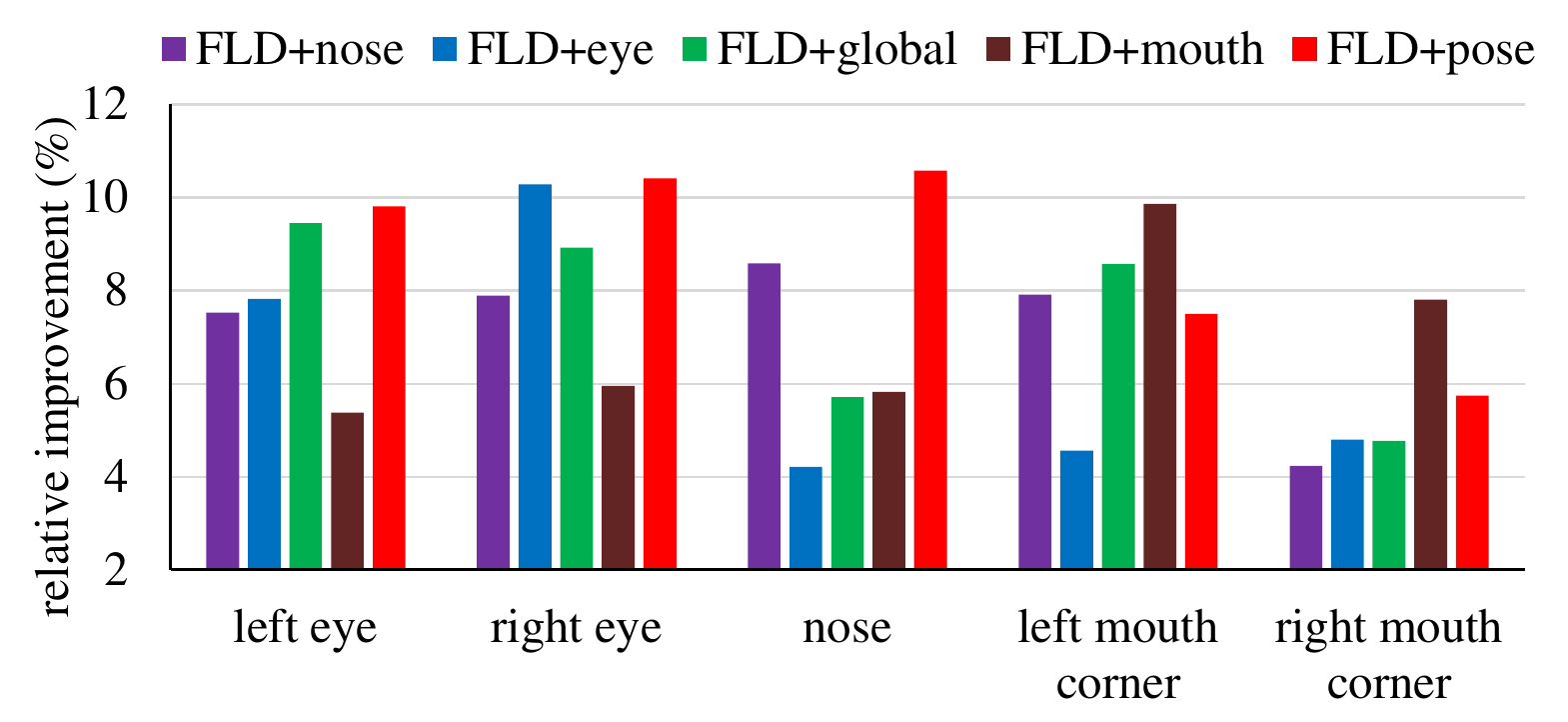}\\
  \vskip -0.3cm
  \caption{Improvement over different landmarks by different attribute groups.}
  \label{fig:model_improvement}

\end{figure}

\begin{figure}
  \centering
  \includegraphics[width=0.4\textwidth]{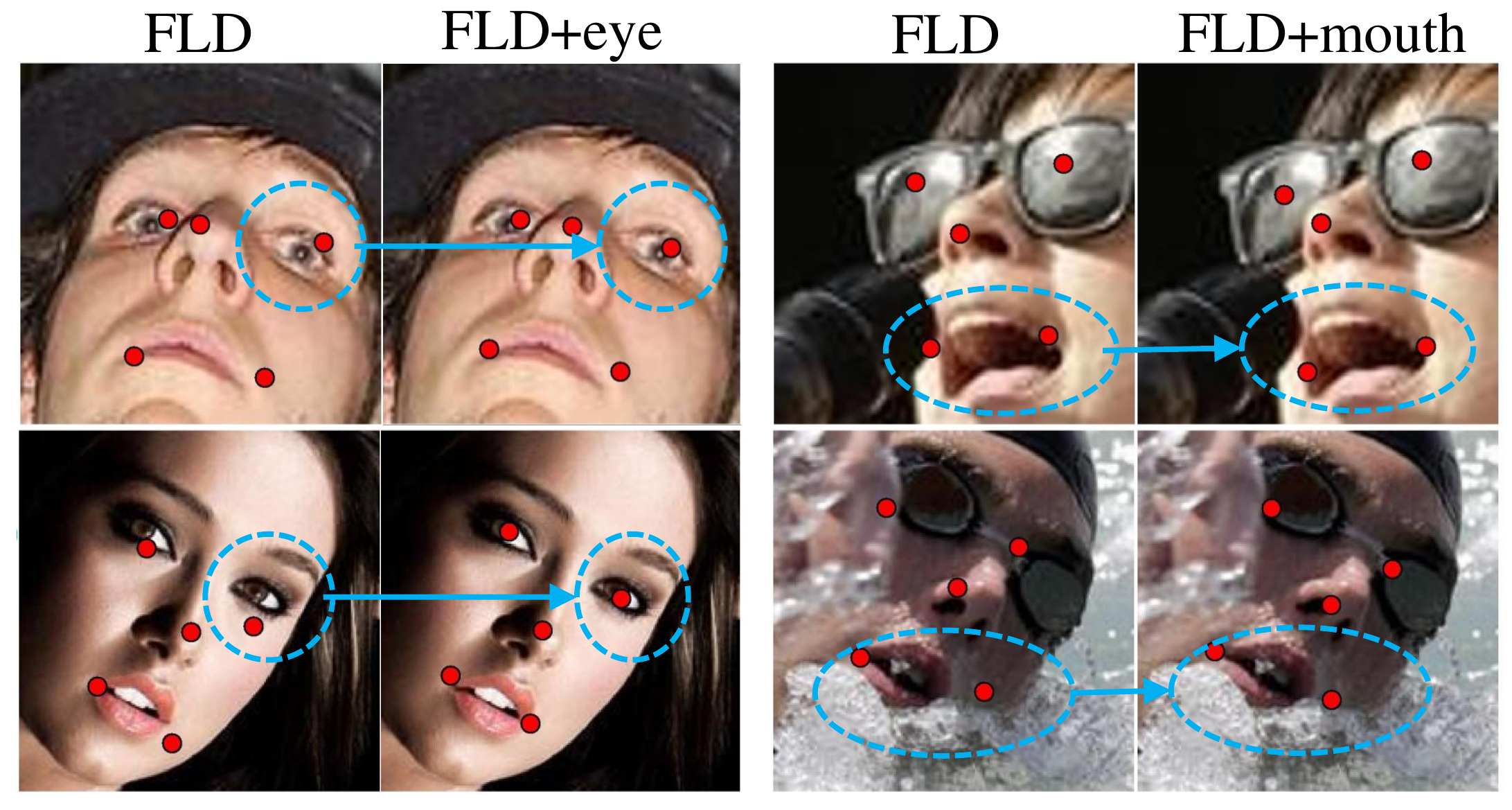}\\
  \vskip -0.3cm
  \caption{Examples of improvement by attribute group of ``eye'' and ``mouth''.}
  \label{fig:vis_improvement}

\end{figure}

In addition, we show the relative improvement caused by different groups of attributes for each landmark in Fig.~\ref{fig:model_improvement}. In particular, we define $\textrm{relative improvement} = \frac{\textrm{reduced error}}{\textrm{original error}}$, where original error is produced by the model of ``FLD only''. We can observe a trend that each group facilitates the landmarks in the according face region. For example, for the group of ``mouth'', the benefits are mainly observed at the corners of mouth. This observation is intuitive since attributes like smiling drives the lower part of the faces, involving Zygomaticus and levator labii superioris muscles, more than the upper facial region. The learning of these attributes develops a shared representation that describes lower facial region, which in turn facilitates the localization of corners of mouth. Similarly, the improvement of eye location is much more significant than mouth and nose for the attribute group of ``eye''. However, we observe the group of ``nose'' improves the eye and mouth localization remarkably. This is mainly because the nose is in the central of the face, there exists constrain between the nose location and other landmarks. The horizontal coordinate of the nose is likely to be the mean of the eyes in frontal face. As for the group of ``pose'' and ``global'', the improvement is significant in all landmarks. Fig.~\ref{fig:vis_improvement} depicts improvements led by adding ``eye'' and ``mouth'' attributes. Fig.~\ref{fig:visual_examples_MTFL_AFLW} shows more example results, demonstrating the effectiveness on various face appearances of TCDCN.


\subsection{Comparison with Deep Learning based Methods}
\label{subset:cmp_deep}
Although the TCDCN, cascaded CNN~\cite{Sun2013} and CFAN~\cite{zhang2014coarse} are built upon deep model, we show that the proposed model can achieve better detection accuracy with lower computational cost and model complexity. We use the full model ``FLD+all'', and the publicly available binary code of cascaded CNN~\cite{Sun2013} and CFAN~\cite{zhang2014coarse} in this experiment.

\begin{figure}[t]
  \centering
  \includegraphics[width=0.48\textwidth]{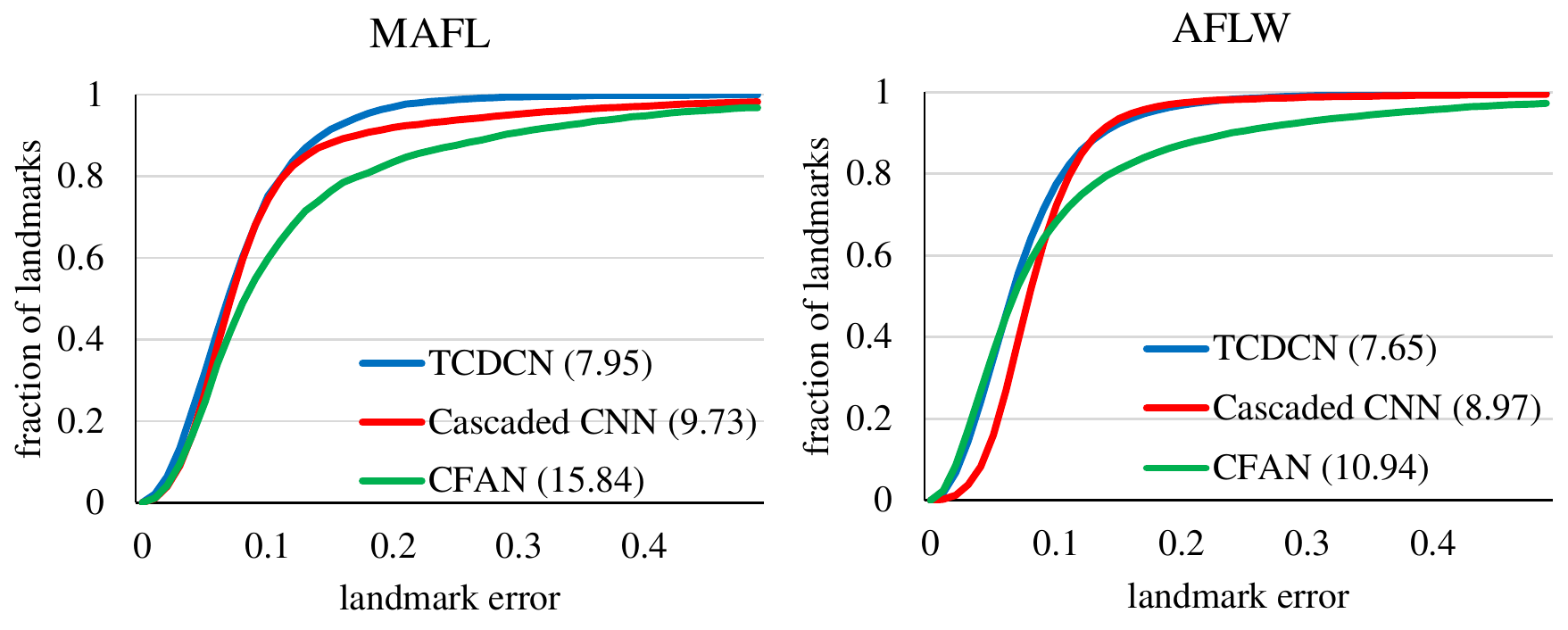}
  \vskip -0.3cm
  \caption{Cumulative error curves of the proposed method, cascaded CNN~\cite{Sun2013} and CFAN~\cite{zhang2014coarse}, on the MAFL and AFLW~\cite{Kostinger2011} dataset (5 landmarks). The number in the legend indicates the according mean error ($\times 10^{-2}$).}
  \label{fig:cmp_deep}
\end{figure}

\noindent\textbf{Landmark localization accuracy:}
In this experiment, we employ the testing images of MAFL and AFLW~\cite{Kostinger2011} for evaluation. It is observed from Fig.~\ref{fig:cmp_deep} that the overall accuracy of the proposed method is superior to that of cascaded CNN and CFAN.

\vspace{0.1cm}
\noindent\textbf{Model complexity}:
The proposed method only has one CNN, whereas the cascaded CNN~\cite{Sun2013} deploys multiple CNNs in different cascaded layers (23 CNNs in its implementation). Also, for each CNN, both our method and cascaded CNN~\cite{Sun2013} have four convolutional layers and two fully connected layers, with comparable input image size. However, the convolutional layer in~\cite{Sun2013} uses locally unshared kernels. Hence, TCDCN has much lower computational cost and model complexity. The cascaded CNN requires 0.12s to process an image on an Intel Core i5 CPU, whilst TCDCN only takes 18ms, which is \textit{7 times faster}. Also, the TCDCN costs 1.5ms on a NVIDIA GTX760 GPU. Similarly, the complexity is larger in CFAN~\cite{zhang2014coarse} due to the use of multiple auto-encoders, each of which contains fully connected structures in all layers. Table~\ref{tab:complexity} shows the details of the running time and network complexity.

\begin{table}[t]
\caption{Comparison of Different Deep Modes for Facial Landmark Detection}
\label{tab:complexity}
\vskip -0.75cm
\begin{center}
\begin{tabular}{l|c|c}
\hline
Model&\#Parameter&Time (per face)\\
\hline\hline
Cascaded CNN~\cite{Sun2013}&$\sim$990K&120ms\\
CFAN~\cite{zhang2014coarse}&$\sim$18,500K&30ms\\
TCDCN&$\sim$100K&18ms\\

\hline
\end{tabular}
\end{center}
\vskip -0.4cm
\end{table}

\vskip -0.3cm
\subsection{Face Representation Transfer and Comparison with State-of-the-art Methods}
\label{subset:cmp_state_of_the_art}
As we discussed in Section~\ref{sec:300w}, we can transfer the trained TCDCN to handle more landmarks beyond the five major facial points.
The main idea is to pre-train a TCDCN on sparse landmark annotations and multiple auxiliary tasks, followed by fine-tuning with dense landmark points.

We compare against various state-of-the-arts. The first class of methods use regression methods that directly predict the facial landmarks: (1) Robust Cascaded Pose Regression (RCPR)~\cite{dollar13}; (2) Explicit Shape Regression (ESR)~\cite{Cao2012}; (3) Supervised Descent Method (SDM)~\cite{6618919}; (4) Regression based on Local Binary Features (LBF)~\cite{300w_lbp}; (5) Regression based on Ensembles of Regression Trees~\cite{kazemi2014one} (ERT); (6) Coarse-to-Fine Auto-Encoder Networks (CFAN)~\cite{zhang2014coarse} (as this method can predict dense landmarks, we compare with it again); (7) Coarse-to-Fine Shape Searching (CFSS)~\cite{zhu2015face}. The second class of methods employ a face template: (8) Tree Structured Part Model (TSPM)~\cite{Zhu2012}, which jointly estimates the head pose and facial landmarks; (9) A Cascaded Deformable Model (CDM)~\cite{Yu2013}; (10) STASM~\cite{milborrow2008locating}, which is based on Active Shape Model~\cite{cootes1995active}; (11) Component-based ASM~\cite{Helen}; (12) Robust discriminative response map fitting (DRMF) method~\cite{asthana2013robust}; (13) Gauss-newton deformable part models (GN-DPM)~\cite{tzimiropoulos2014gauss}; In addition, we compare with the  commercial face analysis software: (14) Face++ API~\cite{FACEPP}. For the methods of RCPR~\cite{dollar13}, SDM~\cite{6618919}, CFAN~\cite{zhang2014coarse}, TSPM~\cite{Zhu2012}, CDM~\cite{Yu2013}, STASM~\cite{milborrow2008locating}, DRMF~\cite{asthana2013robust}, and Face++~\cite{FACEPP}, we use their publicly available implementation. For the methods which include their own face detector (like TSPM~\cite{Zhu2012} and CDM~\cite{Yu2013}), we avoid detection errors by cropping the image around the face. For methods that do not release the code, we report their results on the related literatures.

%

\begin{figure}[t]
  \centering
  \includegraphics[width=0.4\textwidth]{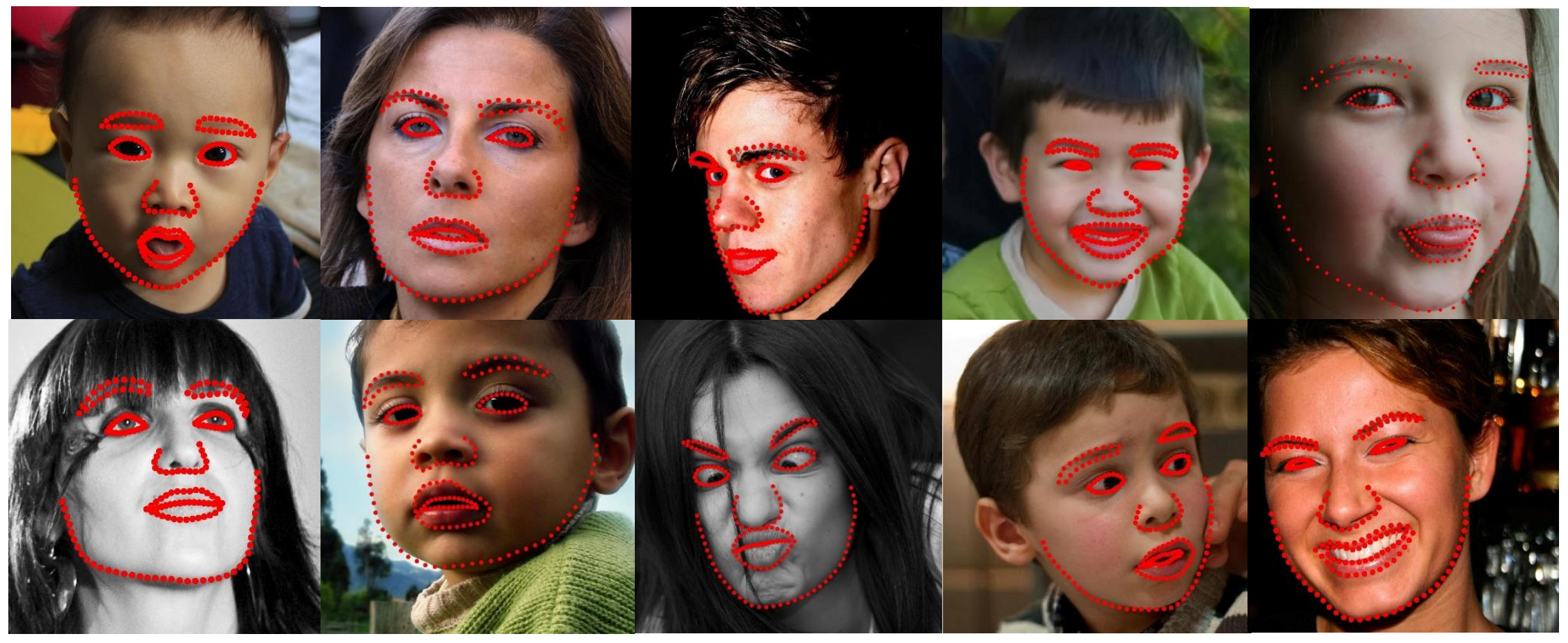}
  \vskip -0.3cm
  \caption{Example alignment results on Helen~\cite{Helen}.}
  \label{fig:visual_examples_helen}
  \vskip -0.3cm
\end{figure}

\vspace{0.1cm}
\noindent\textbf{Evaluation on Helen~\cite{Helen}}: Helen consists of 2,000 training and 330 testing images as specified in~\cite{Helen}. In particular, the 194-landmark annotation is from the original dataset and the 68-landmark annotation is from~\cite{300w}. Table~\ref{tab:helen} reports the performance of the competitors and the proposed method. Most of the images are in high resolution and the faces are near-frontal. Although our method just uses the input of $60\times60$ grey image, it still achieves better result. Fig.~\ref{fig:visual_examples_helen} visualizes some of our results. It can be observed that driven by rich facial attributes, our model can capture various facial expression accurately.

\begin{table}[t]
\caption{Mean errors (\%) on Helen~\cite{Helen} dataset}
\vskip -0.5cm
\label{tab:helen}

\begin{center}
\begin{tabular}{l|c|c}
\hline
Method&194 landmarks&68 landmarks\\
\hline\hline
STASM~\cite{milborrow2008locating}&11.1&-\\
CompASM~\cite{Helen}&9.10&-\\
DRMF~\cite{asthana2013robust}&-&6.70\\
ESR~\cite{Cao2012}&5.70&-\\
RCPR~\cite{dollar13}&6.50&5.93\\
SDM~\cite{6618919}&5.85&5.50\\
LBF~\cite{300w_lbp}&5.41&-\\
CFAN~\cite{zhang2014coarse}&-&5.53\\
CDM~\cite{Yu2013}&-&9.90\\
GN-DPM~\cite{tzimiropoulos2014gauss}&-&5.69\\
ERT~\cite{kazemi2014one}&4.90&-\\
CFSS~\cite{zhu2015face}&4.74&4.63\\
TCDCN&\textbf{4.63}&\textbf{4.60}\\
\hline
\end{tabular}
\end{center}
\end{table}

\begin{figure*}[t]
  \centering
  \includegraphics[width=0.75\textwidth]{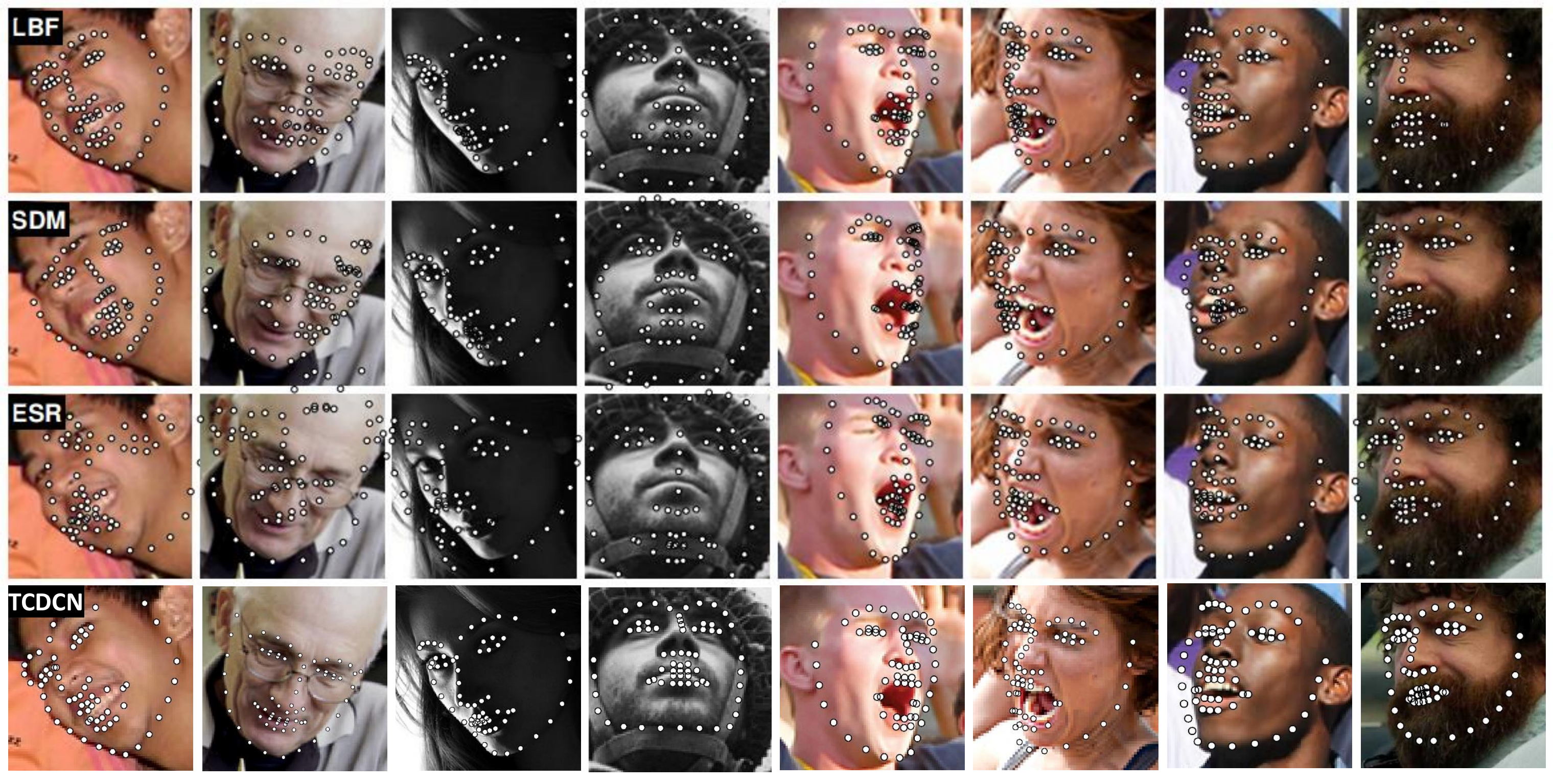}
  \vskip -0.3cm
  \caption{Results of ESR~\cite{Cao2012}, SDM~\cite{6618919}, LBF~\cite{300w_lbp} and our method on the IBUG faces~\cite{300w}.}
  \label{fig:visual_examples_300W}
\end{figure*}

\begin{figure*}[t]
  \centering
  \includegraphics[width=0.75\textwidth]{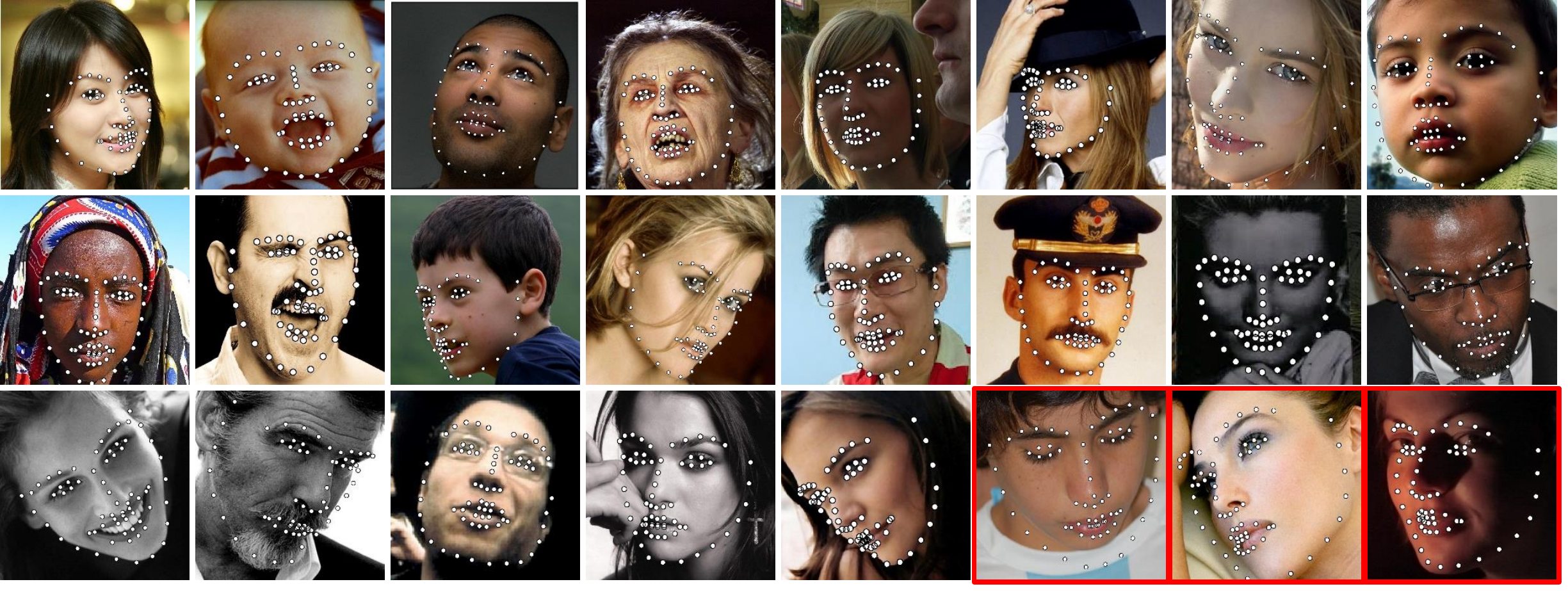}
  \vskip -0.3cm
  \caption{Example alignment results on Helen~\cite{Helen}, IBUG~\cite{300w}, and LPFW~\cite{Belhumeur2011} datasets (68 landmarks). Red rectangles indicate wrong cases.}
  \label{fig:visual_examples_HELEN-IBUG-300W}
\end{figure*}

\vspace{0.1cm}
\noindent\textbf{Evaluation on 300-W~\cite{300w}}: We follow the same protocol in~\cite{300w_lbp}: the training set contains 3,148 faces, including the AFW, the training sets of LFPW, and the training sets of Helen. The test set contains 689 faces, including IBUG, the testing sets of LFPW, and the testing sets of Helen. Table~\ref{tab:300w} demonstrates the superior of the proposed method. In particular, for the challenging subset (IBUG faces) TCDCN produces a significant error reduction of over \textbf{10\%} in comparison to the state-of-the-art~\cite{zhu2015face}. As can be seen from Fig.~\ref{fig:visual_examples_300W}, the proposed method exhibits superior capability of handling difficult cases with large head rotation and exaggerated expressions, thanks to the shared representation learning with auxiliary tasks. Fig.~\ref{fig:visual_examples_HELEN-IBUG-300W} shows more results of the proposed method on Helen~\cite{Helen}, IBUG~\cite{300w}, and LPFW~\cite{Belhumeur2011} datasets.
\begin{table}[t]
\caption{Mean errors (\%) on 300-W~\cite{300w} dataset (68 landmarks)}
\vskip -0.5cm
\label{tab:300w}
\begin{center}
\begin{tabular}{l|c|c|c}
\hline
Method&Common Subset&Challenging Subset&Fullset\\
\hline\hline
CDM~\cite{Yu2013}&10.10&19.54&11.94\\
DRMF~\cite{asthana2013robust}&6.65&19.79&9.22\\
RCPR~\cite{dollar13}&6.18&17.26&8.35\\
GN-DPM~\cite{tzimiropoulos2014gauss}&5.78&-&-\\
CFAN~\cite{zhang2014coarse}&5.50&16.78&7.69\\
ESR~\cite{Cao2012}&5.28&17.00&7.58\\
SDM~\cite{6618919}&5.57&15.40&7.50\\
ERT~\cite{kazemi2014one}&-&-&6.40\\
LBF~\cite{300w_lbp}&4.95&11.98&6.32\\
CFSS~\cite{zhu2015face}&\textbf{4.73}&9.98&5.76\\
TCDCN&4.80&\textbf{8.60}&\textbf{5.54}\\
\hline
\end{tabular}
\end{center}
 \vskip -0.3cm
\end{table}

\begin{figure}[t]
  \centering
  \includegraphics[width=0.3\textwidth]{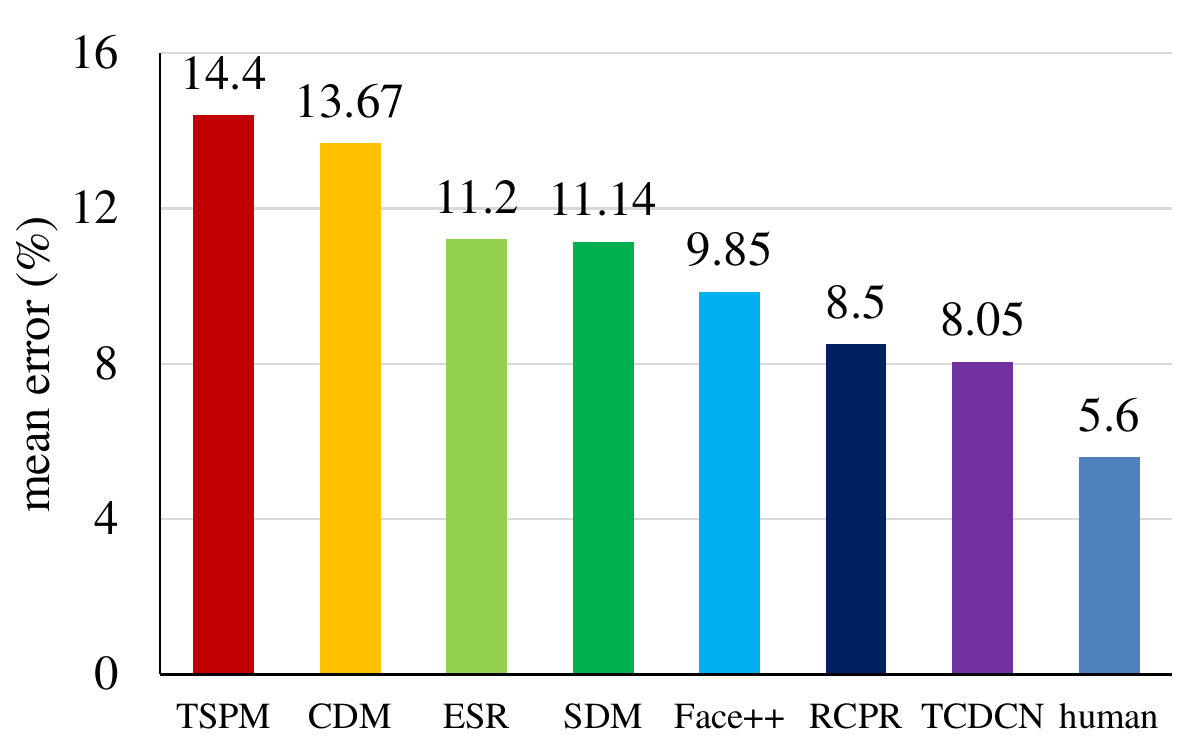}
   \vskip -0.3cm
  \caption{Mean errors on COFW~\cite{dollar13} dataset (29 landmarks) for the method of TSPM~\cite{Zhu2012}, CDM~\cite{Yu2013}, ESR~\cite{Cao2012}, SDM~\cite{6618919}, Face++~\cite{FACEPP}, RCPR~\cite{dollar13} and the proposed method.}
  \label{fig:cmp_cofw}
   \vskip -0.3cm
\end{figure}

\begin{figure}[t]
  \centering
  \includegraphics[width=0.4\textwidth]{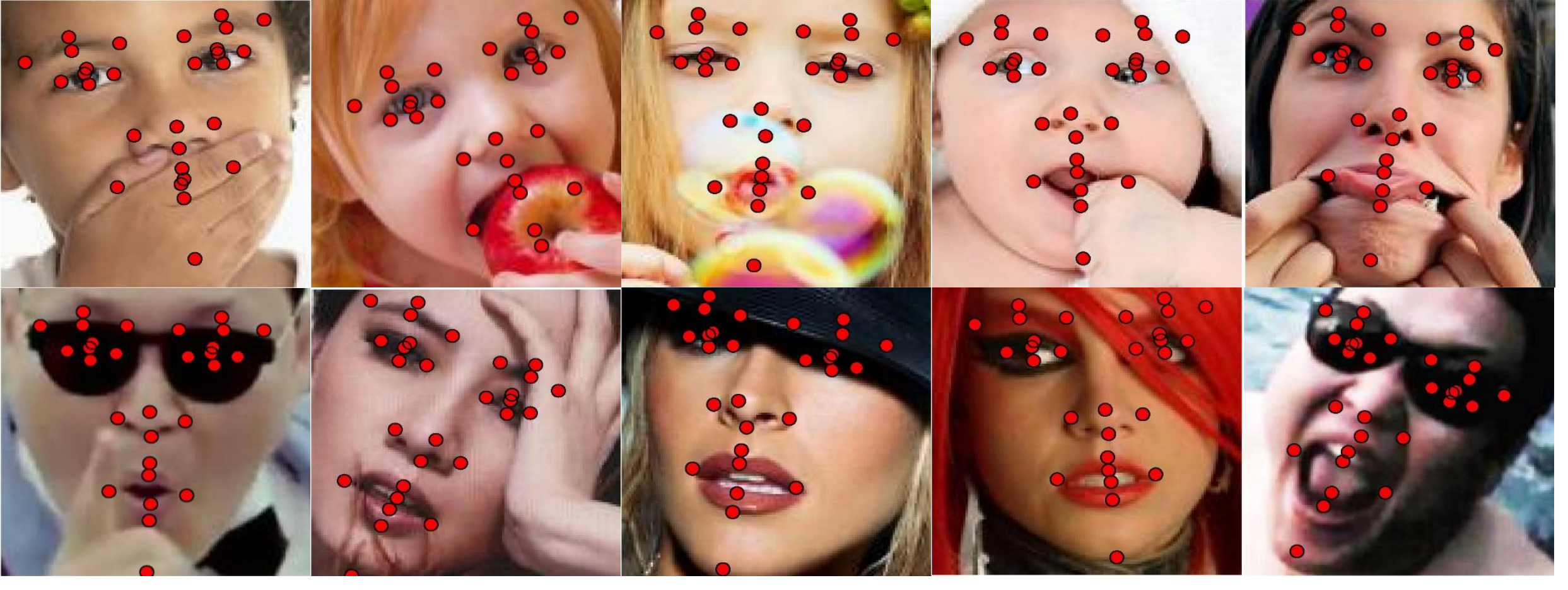}
  \vskip -0.3cm
  \caption{Example alignment results on COFW~\cite{dollar13} dataset.}
  \label{fig:visual_examples_COFW}
  \vskip -0.3cm
\end{figure}

\vspace{0.1cm}
\noindent\textbf{Evaluation on COFW~\cite{dollar13}}: The testing protocol is the same as~\cite{dollar13}, where the training set includes LFPW~\cite{Belhumeur2011} and 500 COFW faces, and the testing set includes the remaining 507 COFW faces. This dataset is more challenging as it is collected to emphasize the occlusion cases. The quantitative evaluation is reported in Fig.~\ref{fig:cmp_cofw}. Example results of our algorithm are depicted in Fig.~\ref{fig:visual_examples_COFW}.
It is worth pointing out that the proposed method achieves better performance than RCPR~\cite{dollar13} even that we do not explicitly learn and detect occlusions as~\cite{dollar13}.

\vskip -0.6cm

\section{Conclusions}
\vskip -0.1cm
Instead of learning facial landmark detection in isolation, we have shown that more robust landmark detection can be achieved through joint learning with heterogeneous but subtly correlated auxiliary tasks, such as appearance attribute, expression, demographic, and head pose. The proposed Tasks-Constrained DCN allows errors of auxiliary tasks to be back-propagated in deep hidden layers for constructing a shared representation to be relevant to the main task. We also show that by learning dynamic task coefficient, we can utilize the auxiliary tasks in a more efficient way. Thanks to learning with the auxiliary attributes, the proposed model is more robust to faces with severe occlusions and large pose variations compared to existing methods. We have observed that a deep model needs not be cascaded~\cite{Sun2013} to achieve the better performance. The lighter-weight CNN allows real-time performance without the usage of GPU or parallel computing techniques. Future work will explore deep learning with auxiliary information for other vision problems.
%
%
%
%



%
\vskip -0.3cm
\bibliographystyle{IEEEtran}
\bibliography{short-PAMI,egbib}

%

\begin{IEEEbiography}[{\includegraphics[width=1in,height=1.25in,clip,keepaspectratio]{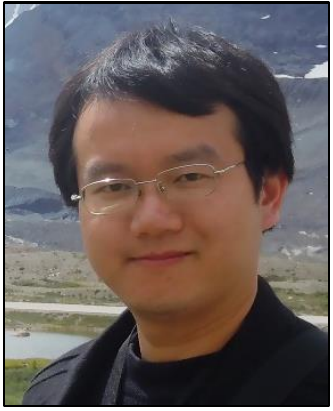}}]{Zhanpeng Zhang}
received the B.E. and M.E. degrees from Sun Yat-sen University, Guangzhou, China, in 2010 and 2013, respectively. He is currently working toward the Ph.D. degree with the Department of Information Engineering, The Chinese University of Hong Kong, Hong Kong. His research interests include computer vision and machine learning, in particular, face tracking and analysis.
\end{IEEEbiography}

\begin{IEEEbiography}[{\includegraphics[width=1in,height=1.25in,clip,keepaspectratio]{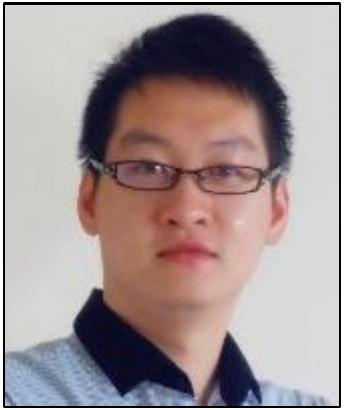}}]{Ping Luo} is a postdoctoral researcher at the Department of Information Engineering, The Chinese University of Hong Kong, where he received his Ph.D. in 2014. His research interests include deep learning, computer vision, and computer graphics, focusing on face analysis, pedestrian analysis, and large-scale object recognition and detection.
\end{IEEEbiography}


\begin{IEEEbiography}[{\includegraphics[width=1in,height=1.25in,clip,keepaspectratio]{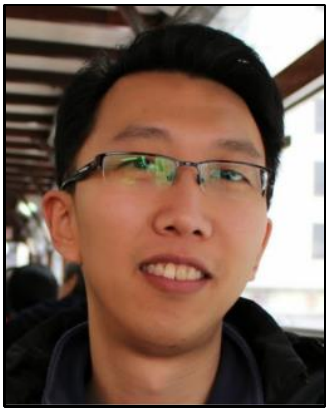}}]{Chen Change Loy}
received the PhD degree in Computer Science from the Queen Mary University of London in 2010. He is currently a Research Assistant Professor in the Department of Information Engineering, Chinese University of Hong Kong. Previously he was a postdoctoral researcher at Vision Semantics Ltd. His research interests include computer vision and pattern recognition, with focus on face analysis, deep learning, and visual surveillance.
\end{IEEEbiography}

\begin{IEEEbiography}[{\includegraphics[width=1in,height=1.25in,clip,keepaspectratio]{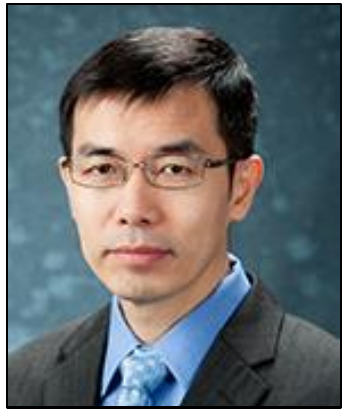}}]{Xiaoou Tang}
received the B.S. degree from the University of Science and Technology of China, Hefei, in 1990, and the  M.S. degree from the University of Rochester, Rochester, NY, in  1991. He received the Ph.D. degree from the Massachusetts Institute of Technology, Cambridge, in 1996. He is a Professor in the Department of Information Engineering and Associate Dean (Research) of the Faculty of Engineering of the Chinese University of Hong Kong. He worked as the group manager of the Visual Computing Group at the Microsoft Research Asia from 2005 to 2008. His research interests include computer vision, pattern recognition, and video processing. Dr. Tang received the Best Paper Award at the IEEE Conference on Computer Vision and Pattern Recognition (CVPR) 2009. He is a program chair of the IEEE International Conference on Computer Vision (ICCV) 2009 and has served as an Associate Editor of IEEE Transactions on Pattern Analysis and Machine Intelligence (PAMI) and International Journal of Computer Vision (IJCV). He is a Fellow of IEEE.
\end{IEEEbiography}

\end{document}